\documentclass[lettersize,journal]{IEEEtran}
\usepackage[
  colorlinks=true,
  linkcolor=magenta,
  urlcolor=magenta
]{hyperref}
\usepackage{amsmath,amsfonts}
\usepackage{algorithmic}
\usepackage{algorithm}
\usepackage{array}
\usepackage[caption=false,font=normalsize,labelfont=sf,textfont=sf]{subfig}
\usepackage{textcomp}
\usepackage{stfloats}
\usepackage{url}
\usepackage{verbatim}
\usepackage{amssymb}
\usepackage{dsfont}
\usepackage{graphicx}
\usepackage{cite}
\usepackage{multirow}
\usepackage{booktabs}
\usepackage{siunitx}
\usepackage{xcolor}
\usepackage{amsmath}
\usepackage[switch, modulo]{lineno}
\usepackage{nomencl}

\usepackage{imakeidx} 
\makenomenclature
\makeindex
\begin{document}

\title{Quality-focused Active Adversarial Policy for Safe Grasping in Human-Robot Interaction}
\author{Chenghao Li, Razvan Beuran, \IEEEmembership{Senior Member, IEEE}, and Nak Young Chong, \IEEEmembership{Senior Member, IEEE}
\thanks{This work was supported by JSPS KAKENHI Grant Number JP23K03756 and the Asian Office of Aerospace Research and Development under Grant/Cooperative Agreement Award No. FA2386-22-1-4042.}
\thanks{The authors are with the School of Information Science, Japan Advanced Institute of Science and Technology, 1-1, Asahidai, Nomi, 923-1292, Ishikawa, Japan (e-mail: chenghao.li@jaist.ac.jp; nakyoung@jaist.ac.jp).}}

\markboth{Preprint Version}%
{Shell \MakeLowercase{\textit{et al.}}: A Sample Article Using IEEEtran.cls for IEEE Journals}


\maketitle

\begin{abstract}
Vision-guided robot grasping methods based on Deep Neural Networks (DNNs) have achieved remarkable success in handling unknown objects, attributable to their powerful generalizability. However, these methods with this generalizability tend to recognize the human hand and its adjacent objects as graspable targets, compromising safety during Human-Robot Interaction (HRI). In this work, we propose the Quality-focused Active Adversarial Policy (QFAAP) to solve this problem. Specifically, the first part is the Adversarial Quality Patch (AQP), wherein we design the adversarial quality patch loss and leverage the grasp dataset to optimize a patch with high quality scores. Next, we construct the Projected Quality Gradient Descent (PQGD) and integrate it with the AQP, which contains only the hand region within each real-time frame, endowing the AQP with fast adaptability to the human hand shape. Through AQP and PQGD, the hand can be actively adversarial with the surrounding objects, lowering their quality scores. Therefore, further setting the quality score of the hand to zero will reduce the grasping priority of both the hand and its adjacent objects, enabling the robot to grasp other objects away from the hand without emergency stops. We conduct extensive experiments on the benchmark datasets and a cobot, showing the effectiveness of QFAAP. Our code and demo videos are available here: 
\href{https://github.com/clee-jaist/QFAAP}{\url{https://github.com/clee-jaist/QFAAP}}
\end{abstract}


\section{INTRODUCTION}

\begin{figure}[!t]
\vspace{0.3\baselineskip}
\centerline{\includegraphics[width=\columnwidth]{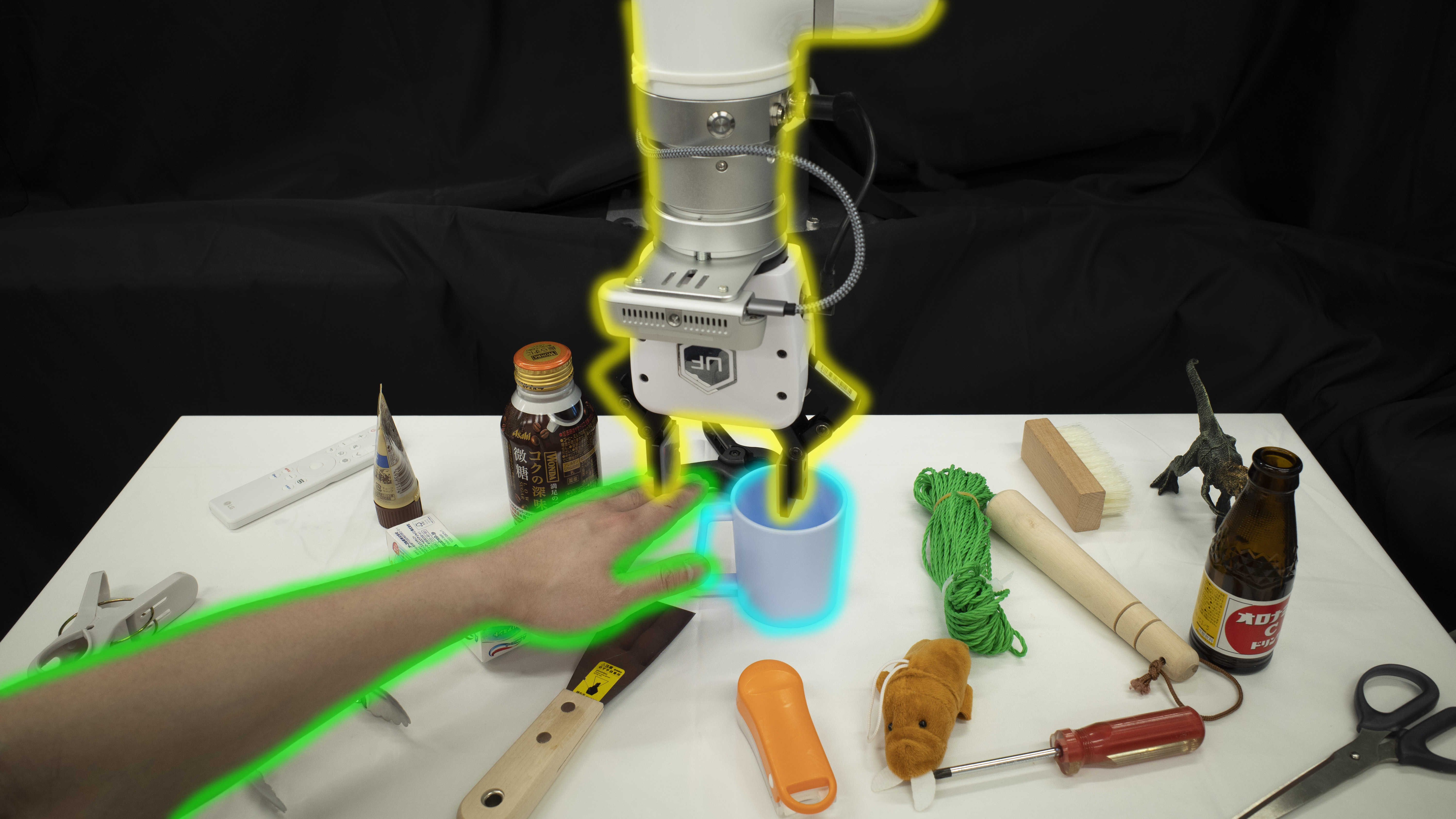}}
\caption{An example of a cluttered HRI scenario: the robot mistakenly identifies the human hand or adjacent objects as graspable targets for autonomous grasping, causing harm to the human. We highlight the robot, the human hand, and the target object using yellow, green, and blue borders, respectively.}
\label{fig1}
\end{figure}

\IEEEPARstart{V}{ISION}-guided robot grasping is one of the critical capabilities for HRI \cite{c1}, aimed at helping humans improve work efficiency in the service and manufacturing domain. Traditional visual grasping methods typically construct a grasp database based on three-dimensional (3D) object models, incorporating performance metrics derived from geometric and physical properties \cite{c1, c2} and employing stochastic sampling to account for grasping uncertainty \cite{c3}. However, these methods are inherently limited by their reliance on known 3D object models, rendering them ineffective when applied to novel objects. To address this limitation, recent studies \cite{c4, c5} have introduced an alternative paradigm that leverages DNNs \cite{c6, c7, c8, c9, c10} to train function approximators. These approximators predict grasp candidates directly from images, utilizing datasets comprising empirical grasp successes and failures, thereby enabling efficient generalization to previously unseen objects at substantially lower cost. However, these methods with this generalizability may also recognize the human hand and its adjacent objects as graspable targets, compromising safety during HRI, as shown in Fig. \ref{fig1}. Given the growing trend of large-scale deployment of DNNs-based visual grasping systems in HRI scenarios, ignoring this safety issue could lead to workplace injuries and accidents. 

Some methods assist robots in avoiding 
collisions with human hands and enabling interaction by segmenting human hands or estimating their pose or motion, as exemplified in Robot-to-Human Handover (R2H) \cite{c11} and Human-to-Robot Handover (H2R) \cite{c12, c13, c14, c15}. Although these methods are effective in helping robots avoid human hands during HRI, most focus on the handover problem between humans and robots in simple single-object scenarios. In contrast, this paper will emphasize the problem of enabling robots to autonomously avoid the human hand and objects close to the hand for grasping operations without emergency stops in complex cluttered HRI scenarios, which is a new and more challenging problem in DNNs-based visual grasping.

How to address this problem? Inspired by adversarial attacks~\cite{c16, c17, c18}, which leverage the interpretability flaws of DNNs to craft perturbations that interfere with model predictions, we investigate from a novel perspective: whether adversarial attacks can be used as benign adversarial perturbations to interfere with the grasp quality score, thereby dynamically adjusting the grasping sequence of the robot to actively avoid the human hand and objects adjacent to it. Therefore, based on this new perspective, the method we aim to design differs significantly from common adversarial attacks. Firstly, most adversarial attack methods focus on how to attack the model. In contrast, our goal is not to attack or defend \cite{c19}  but to address the safety issue in DNNs-based visual grasping within HRI scenarios through controllable perturbations. Secondly, our method emphasizes actively perturbing the grasp quality score to alter the grasping priority of human hands and their neighboring objects, thereby guiding the robot to avoid grasping them. In contrast, common adversarial attacks primarily aim to degrade detection accuracy~\cite{c20, c21}, cause misclassification \cite{c22, c23} or mislocalization \cite{c21}, and evade detection~\cite{c24, c25, c26}. Finally, since human hands can appear with arbitrary postures to perform tasks in various HRI scenarios, the perturbation we want to design must conform closely to the shape of the hand at a breakneck speed, keeping the hand away from the robot gripper. This is much more difficult than other adversarial attacks \cite{c16, c20, c21} that apply perturbations with fixed shapes or extend to other specific shapes through complicated processes and high costs \cite{c25, c26, c27}.

Along these lines, this paper proposes the Quality-focused Active Adversarial Policy (QFAAP), which first optimizes an Adversarial Quality Patch (AQP) with high quality scores by the adversarial quality patch loss and grasp dataset. Next, integrate AQP that contains only the hand region within each real-time frame with the Projected Quality Gradient Descent (PQGD), ensuring AQP has fast adaptability to the human hand shape. By applying AQP and PQGD, the hand can actively interfere with nearby objects, reducing their quality score. Further, setting the quality score of the hand to zero will simultaneously lower the grasping priority of both the hand and surrounding objects, enabling the robot to actively avoid them while grasping without emergency stops.

A summary of the contributions in this work is as follows:

\begin{enumerate}
\item We reveal a new and more challenging problem in DNNs-based visual grasping: how to enable robots to simultaneously avoid human hands and nearby objects without emergency stops during grasping in cluttered HRI scenarios.
\item We propose a novel safety-oriented grasping policy from a benign adversarial perspective, named QFAAP, which can actively perturb the grasp quality score to alter the grasping priority, thereby enabling robots to avoid human hands and nearby objects during grasping. To the best of our knowledge, this is the first study on benign adversarial in real robot grasping.
\item We validate the effectiveness of our proposed method through comprehensive experiments on three benchmark datasets and a real cobot across various single-object and clutter scenarios.
\end{enumerate}

This paper is organized into the following sections. Section II (Related Work) reviews vision-guided robot grasping and adversarial attacks. Section III (Proposed Method) provides an overview of QFAAP, detailing its two components (AQP and PQGD), and discusses how QFAAP is implemented in robot grasping. Section IV (Experiments) validates the effectiveness of our method in benchmark datasets and real-world grasping scenarios. Finally, Section V (Conclusion) summarizes the work of this paper and provides prospects for future research.

\section{Related Work}

\subsection{Vision-guided Robot Grasping}
While many grasping frameworks exist, this work focuses explicitly on vision-guided 4-Degree-of-Freedom (4-DOF) grasping using a parallel-jaw gripper, which can be broadly categorized into traditional methods and DNNs-based methods. Traditional grasping methods are founded on mathematical and physical models that characterize object geometry, kinematics, and dynamics \cite{c1, c2, c3}. These methods typically assume the availability of a detailed 3D model of the object, which is leveraged to compute stable grasp configurations. For instance, Gallegos {\it et al.} \cite{c28} optimized grasping strategies by utilizing predefined contact points on known 3D object models. Similarly, Pokorny {\it et al.} \cite{c29} introduced the concept of grasping spaces, enabling the mapping of objects to these spaces for grasp synthesis. While these approaches exhibit robustness in structured environments, their applicability is inherently constrained by the prerequisite of complete 3D object models, and they are often unavailable in unstructured environments containing novel objects. This limitation underscores the need for more flexible grasping strategies to handle object uncertainty in unstructured environments.

DNNs-based visual grasping methods demonstrate strong generalization capabilities to novel objects by employing function approximators trained on extensive datasets to predict the grasp success probability from images. Consequently, datasets play a pivotal role in these methods. A notable human-labeled dataset is the Cornell Grasping Dataset \cite{c30}, which comprises approximately 1,000 RGB-D images and has been widely utilized for training grasping models in single-object scenarios~\cite{c31, c32, c33, c34, c35, c36, c37}. The Dex-Net series \cite{c4, c38, c39, c40, c41} introduced a large-scale synthetic dataset that integrates various cluttered environments to acquire cluttered grasping capabilities, significantly advancing the field of visual grasping. Similarly, GraspNet \cite{c5, c42, c43} constructed a real-world dataset encompassing one billion grasp labels and nearly 100,000 images across 190 densely cluttered scenes and support both 4-DOF and 6-Degree-of-Freedom (6-DOF) grasping, which further improves the grasping ability for unknown objects in cluttered scenarios.

Although the aforementioned DNNs-based methods demonstrate strong generalization capabilities for unknown objects in unstructured environments, they emphasize grasp generalization while neglecting grasp safety. Specifically, these methods with this generalizability will also recognize human hands and adjacent objects as graspable targets, compromising safety during HRI.

\subsection{Adversarial Attacks}
Since Szegedy {\it et al.}. \cite{c44} first identified adversarial examples, extensive research has been conducted to expose the vulnerability of DNNs. These efforts generally fall into two categories: single-image adversarial attacks and image-agnostic attacks (adversarial patch attacks). Single-image adversarial attacks achieve their attacks by maximizing the discriminative loss of the model to generate global perturbations that cover the entire image. Goodfellow {\it et al.}. \cite{c16} designed a Fast Gradient Sign Method (FGSM) to produce strong perturbations based on investigating the model’s linear nature. Wang {\it et al.}. \cite{c45} and Madry {\it et al.}. \cite{c22} further broke the one-step generation of perturbation in FGSM into iterative generation and proposed I-FGSM and Projected Gradient Descent (PGD) attack. Although the single-image adversarial attacks can rapidly attack image classification models, causing them to produce misclassification results, they were limited to one specific image and entire image regions, which means each new image requires re-optimization. Thus, this limitation highlights the need for more flexible methods to attack arbitrary images and any local regions within an image.

Adversarial patch attacks, characterized by their locality and image-agnostic nature, effectively compromise object detection models with localization properties. For instance, Liu {\it et al.}. \cite{c46} designed DPatch to attack widely used object detectors, degrading their detection accuracy and thereby causing mislocalization or misclassification. Later, Lee {\it et al.}. \cite{c47} investigated failure cases of DPatch and subsequently introduced the Robust DPatch. Beyond causing mislocalization or misclassification, some studies focused on evading detection, preventing detectors from recognizing objects occluded by adversarial patches, as explored in \cite{c21, c24}. Later works, such as \cite{c25, c26, c27}, extended adversarial patches by replicating them into adversarial clothing, enabling more flexible evasion across different viewing angles. However, this replication-based extension is costly and typically limited to the fold variations of clothes.

Overall, the aforementioned single-image adversarial and adversarial patch attacks have demonstrated effectiveness, but how to transform these attacks into controllable benign adversarial to address safety concerns in DNNs-based grasping remains unexplored. Moreover, another important yet underexplored direction is how to actively manipulate the grasp quality score in DNNs-based grasping to alter the grasping priority of the robot. Finally, rapidly achieving shape adaptability for adversarial perturbations at minimal cost is critical and practical in robot grasping, which often needs to deal with objects with different shapes. So, in this work, we leverage the advantages of single-image adversarial and adversarial patch attacks, and propose a novel active adversarial method with rapid human hand shape adaptability by manipulating the grasp quality score, which aims to address the safety problem of DNNs-based grasping in the HRI process.

\section{Proposed Method}
In this section, we will first make an overview of QFAAP. Then, a comprehensive description of two important modules (AQP and PQGD) will be provided. Finally, we will explain how to deploy QFAAP to improve visual grasping safety in cluttered HRI scenarios.

\subsection{Overview of QFAAP}
\begin{figure*}[!t]
\centerline{\includegraphics[width=\textwidth]{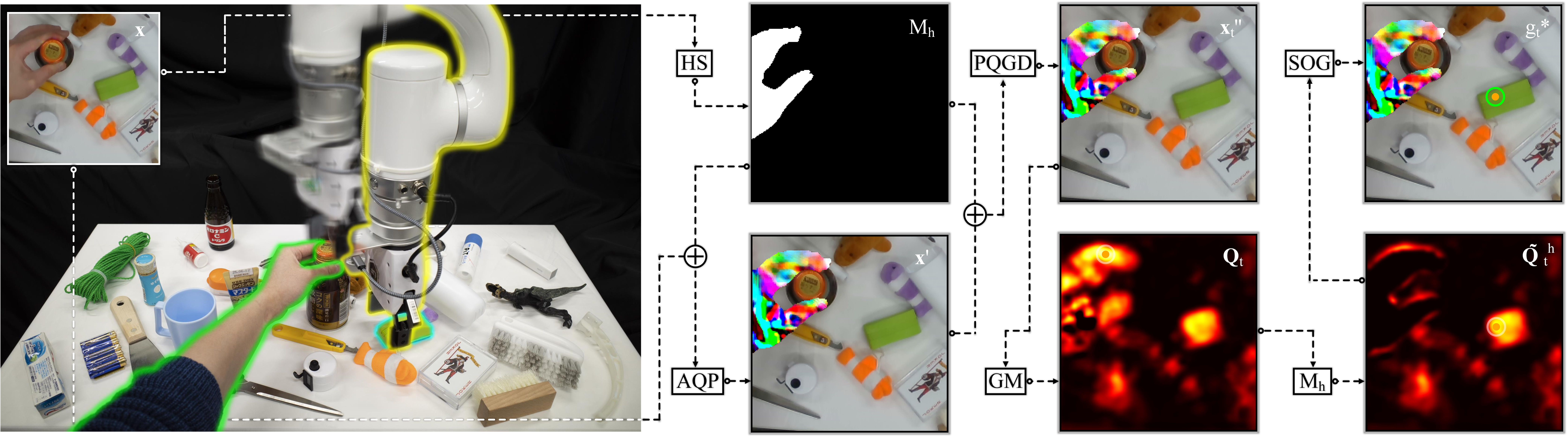}}
\caption{Pipeline of QFAAP: Firstly, the original RGB frame $\mathbf{x}$ is captured by the depth camera, and a hand segmentation algorithm (HS) is applied to obtain the hand mask $\mathcal{M}_h$. Next, the optimized AQP is incorporated into $\mathbf{x}$ while preserving only the hand region, generating $\mathbf{x}'$. In the third stage, PQGD is applied to $\mathbf{x}'$ with $\mathcal{M}_h$ to rapidly endorse the shape adaptability of AQP, producing $\mathbf{x}_t''$. In the fourth stage, $\mathbf{x}_t''$ is fed into the grasping model (GM) to obtain the quality map $\mathbf{Q}_t$, followed by getting the quality map $\tilde{\mathbf{Q}}_t^h$ outside the hand region by $\mathcal{M}_h$. Finally, selecting the optimal grasp (SOG) $g_t^*$ (emphasized by the green circle and orange dot) with the maximum quality score (emphasized by the white circle and orange dot) within $\tilde{\mathbf{Q}}_t^h$. The above process can effectively shift the initial hazardous grasp (the robot is emphasized as a blurred version) located near the hand (emphasized by the green line) toward a safer grasp (the object being grasped and the robot are emphasized with the blue and yellow borders.}
\label{fig2}
\end{figure*}

We propose the Quality-focused Active Adversarial Policy (QFAAP) to enhance the safety of DNNs-based visual grasping in cluttered HRI scenarios. QFAAP consists of two key modules: the Adversarial Quality Patch (AQP) and Projected Quality Gradient Descent (PQGD). The AQP is optimized by the adversarial quality patch loss and grasp dataset, ensuring adversarial effectiveness against the quality score of any image. The PQGD can be integrated with AQP, which contains only the hand region within each real-time frame, endowing AQP with fast human hand shape adaptability. By applying AQP and PQGD, the hand can actively perturb nearby objects to reduce their quality score in the model prediction process. Further, setting the quality score of the hand to zero will simultaneously lower the grasping priority of both the hand and surrounding objects, enabling the robot to actively avoid them while grasping without emergency stops in cluttered HRI scenarios. The pipeline of QFAAP is illustrated in Fig. \ref{fig2}.

\subsection{Adversarial Quality Patch (AQP)}

The DNNs-based visual grasping model typically first defines the grasp configuration \cite{c48}, which is composed of parameters $(j^g, k^g, w^g, h^g, \theta^g)$ forming a rotated box in the image coordinate system, and this box is denoted by the grasp candidate $g_i$. Here, $(j^g, k^g)$ represents the center position of the box, $w^g$ and $h^g$ denote the width and height of the box, and $\theta^g$ represents the angle of the box relative to the horizontal direction. Accordingly, in the robot coordinate system, the grasp and its corresponding parameters are defined as ${\mathcal{G}_i}$ and $(I^g, J^g, Z^g, W^g, \Theta^g)$ (the coordinate transformation from $g_i$ to ${\mathcal{G}_i}$ is explained in Section. \ref{sec:AA}). Then, based on the grasp configuration in the image coordinate system, corresponding objective loss functions are designed, such as the quality loss $\mathcal{L}_q$ associated with $(j^g, k^g)$, the width loss $\mathcal{L}_w$ associated with $w^g$, and the angle loss $\mathcal{L}_\theta$ associated with $\theta^g$. Assuming that for an image sample $x_i$ within one batch (batch size is $B$), the predicted and labeled quality scores at position $n$ of $x_i$ are denoted as ${q}_i(n)$ and $\hat{q}_i(n)$. The quality loss at $n$ of $x_i$ for the model can be defined as Eq. \ref{eq:1}.

\begin{equation} \label{eq:1}
\mathcal{L}_q(n) = 
\begin{cases}
  0.5[q_i(n)- \hat{q}_i(n)], & \text{if} \left|q_i(n)- \hat{q}_i(n)\right| < 1 \\
  \left|q_i(n)- \hat{q}_i(n)\right| - 0.5, & \text{otherwise}
\end{cases}
\end{equation}
By computing the average $\mathcal{L}_q(n)$ across all positions ${N}$, the complete quality loss for the model can be given by Eq. \ref{eq:2}.

\begin{equation} \label{eq:2}
\mathcal{L}_q = \frac{1}{N}\sum\limits_{n=1}^{N} {L}_q(n)
\end{equation}
The losses $\mathcal{L}_w$ and $\mathcal{L}_\theta$ follow the same computation as $\mathcal{L}_q$, consistent with the formulations in Eq. \ref{eq:1} and Eq. \ref{eq:2}. By summing these losses, the total loss for the model can be shown as Eq. \ref{eq:3}.

\begin{equation} \label{eq:3}
\mathcal{L}_{model} = \mathcal{L}_q + \mathcal{L}_\theta + \mathcal{L}_w
\end{equation}

Finally, $\mathcal{L}_{model}$ can be used for model training, where the model weights are optimized via gradient descent. The weight update process is expressed as Eq. \ref{eq:4}. Here, $\mathbf{w}_t$ and $\mathbf{w}_{t-1}$ represent the model weights at time steps $t$ and $t-1$, respectively, while the derivative of $\mathcal{L}_{model}$ with respect to $\mathbf{w}_{t-1}$ denotes the gradient and $\delta_{model}$ is the learning rate of the model. Notably, during training, the quality score within the central one-third region of the grasp label is set to 1 (Maximum), while all other positions are set to 0 (Minimum). This design encourages the model to focus more on learning features in these key regions, thereby increasing the predicted quality score when encountering similar features during inference. Therefore, the quality score is of utmost importance, as it not only determines the grasping position parameters and other parameters corresponding to it, but also dictates the grasping priority, with a higher quality score indicating a higher priority in the grasping sequence.

\begin{equation} \label{eq:4}
\mathbf{w}_t = \mathbf{w}_{t-1} - \delta_{model}\frac{\partial\mathcal{L}_{model}}{\partial\mathbf{w}_{t-1}}
\end{equation}

The AQP is also optimized from the perspective of the quality score. However, unlike optimizing the grasping model, we aim for AQP to optimize in the direction of increasing the quality score rather than minimizing the difference between the predicted quality score and the labeled quality score. Therefore, we first initialize AQP following a uniform distribution, with the same shape as the input image of the model. In optimization, the AQP will be randomly scaled to be applied to the image sample.

Next, we define the quality loss of AQP ($\mathcal{L}_q^p$). let the quality map predicted by the frozen grasping model within the AQP area of $x_i$ be represented as $\mathcal{Q}_i^p$. The quality loss $\mathcal{L}_q^p$ is then defined as in Eq. \ref{eq:5}, where $\mathbb{E}(\mathcal{Q}_i^p)$ and $\text{Var}(\mathcal{Q}_i^p)$ denote the mean and variance of $\mathcal{Q}_i^p$, respectively. The $\alpha$ is an empirical parameter that controls the influence of variance on $\mathcal{L}_q^p$. This loss can be minimized using a gradient descent algorithm by continuously decreasing the negative value (increasing in the negative direction) of $\mathbb{E}(\mathcal{Q}_i^p)$, thereby enhancing the quality score of AQP. So, this can be regarded as the reverse operation of a gradient descent algorithm, achieving gradient ascent to optimize AQP. Additionally, reducing $\text{Var}(\mathcal{Q}_i^p)$ ensures a more stable increase in the quality score.

\begin{equation} \label{eq:5}
\mathcal{L}_q^p = \frac{1}{B}\sum\limits_{i=1}^{B}[-\mathbb{E}(\mathcal{Q}_i^p)+\alpha\text{Var}(\mathcal{Q}_i^p)]
\end{equation}

In this step, we employ the same total variation loss $\mathcal{L}_{tv}$ from \cite{c24} to mitigate noise introduced during AQP optimization, ensuring a smoother optimization, as shown in Eq. \ref{eq:6}. Here, $\mathbf{p}_t(j^p, k^p)$ represents the pixel value of AQP ($\mathbf{p}_t$) at location $(j^p, k^p)$, $W$ and $H$ are the width and height of $\mathbf{p}_t$. This loss is computed as the mean of the Euclidean distance between all adjacent pixel values within AQP.

\begin{equation} \label{eq:6}
\mathcal{L}_{tv} = \frac{1}{H \times W} \sum_{j^p=1}^{H} \sum_{k^p=1}^{W} \left\| \mathbf{p}_t(j^p, k^p) \right\|_2
\end{equation}

To further reinforce the optimization of the quality score for AQP, we introduce the difference loss $\mathcal{L}_{d}$. Let the quality map predicted by the frozen grasping model outside the AQP area of $x_i$ be denoted as $\mathcal{\tilde{Q}}_i^p$. The $\mathcal{L}_{d}$ is defined as in Eq. \ref{eq:7}. This loss can strengthen AQP by letting $\min\mathcal{Q}_i^p$ approach $\max\mathcal{\tilde{Q}}_i^p$. Consequently, AQP will be optimized so that the model predicts a higher quality score for AQP than for other objects in the scene. Thereby, the AQP can effectively interfere with the quality scores of other objects.

\begin{equation} \label{eq:7}
\mathcal{L}_{d} = \frac{1}{B} \sum_{i=1}^{B} \left|\min\mathcal{Q}_i^p-\max\mathcal{\tilde{Q}}_i^p\right|
\end{equation}

Finally, we combine the three aforementioned losses with two additional empirically determined parameters, $\beta$ and $\gamma$, controlling $\mathcal{L}_{tv}$ and $\mathcal{L}_{d}$, respectively, to obtain the total loss of AQP ($\mathcal{L}_{aqp}$), as defined in Eq. \ref{eq:8}. Similarly, we optimize AQP by minimizing this loss using the gradient descent algorithm with Adam optimizer \cite{c49}, as shown in Eq. \ref{eq:9}. Here, $\mathbf{p}_t$ and $\mathbf{p}_{t-1}$ represent AQP at time steps $t$ and $t-1$, respectively, while the derivative of $\mathcal{L}_{aqp}$ with respect to $\mathbf{p}_{t-1}$ denotes the gradient, and $\delta_{aqp}$ is the learning rate of AQP. Since the optimization process is based on the entire grasp dataset, the optimized AQP can be effective on any image.

\begin{equation} \label{eq:8}
\mathcal{L}_{aqp} = \mathcal{L}_q^p + \beta\mathcal{L}_{tv} + \gamma\mathcal{L}_{d}
\end{equation}

\begin{equation} \label{eq:9}
\mathbf{p}_t = \mathbf{p}_{t-1} - \delta_{aqp}\frac{\partial\mathcal{L}_{aqp}}{\partial\mathbf{p}_{t-1}}
\end{equation}

Following the optimized AQP ($\mathbf{p}_t$), we define an evaluation method to assess the quality score level of AQP in one testing batch. Let $j_i^p, k_i^p$ denote the pixel position of the scaled AQP in $x_i$, and let $W_i^p$ and $H_i^p$ as the width and height of the scaled AQP. We compute the ratio $\mathcal{R}_q$ as the proportion of pixels within all AQP regions across a batch where the quality score $\mathcal{Q}_i^p(j_i^p, k_i^p)$ exceeds 0.5, relative to the total number of pixels ($N^p$) in all sample image, as shown in Eq. \ref{eq:10}. Here, $\mathds{1}$ means the indicator function. After defining $\mathcal{R}_q$, we compute the average $\mathcal{R}_q$ for each batch to evaluate the quality score level of AQP across the entire test set, which is denoted by Quality Accuracy (Q-ACC) and will be used in the Experiments section.

\begin{equation} \label{eq:10}
\mathcal{R}_{q} = \frac{1}{N^p} \sum_{i=1}^{B} \left\{\sum_{j_i^p=1}^{H_i^p} \sum_{k_i^p=1}^{W_i^p} \mathds{1} [\mathcal{Q}_i^p(j_i^p, k_i^p) > 0.5]\right\}
\end{equation}

\subsection{Projected Quality Gradient Descent (PQGD)} \label{sec:PQGD}

The PGD \cite{c22} is typically used to attack classification models by inducing misclassification, with the attack targeting the entire region of a single image. In contrast, PQGD primarily focuses on specific local regions within a single image and emphasizes quality score optimization like AQP. Since PQGD, like PGD, exhibits fast optimization properties, it can be employed to further enhance the quality score of local regions in AQP, thereby rapidly endowing AQP with shape adaptability.

Let $\mathbf{x}$ denote a real-time RGB frame from a depth camera, and let $\mathcal{M}_h$ represent the mask of the hand associated with $\mathbf{x}$, obtained using the upper limb segmentation algorithm \cite{c50}. We first define $\mathbf{x'}$ as the RGB frame after adding AQP (the same size as $\mathbf{x}$) within the hand area, as shown in Eq. \ref{eq:11}.

\begin{equation} \label{eq:11}
\mathbf{x}' = \mathbf{x}(1-\mathcal{M}_h) + \mathbf{p}_t\mathcal{M}_h
\end{equation}

Then, let the RGB frame after adding both AQP and PQGD within the hand area be denoted as $\mathbf{x}_t''$. We define the loss of PQGD as $\mathcal{L}_{pqgd}$, as shown in Eq. \ref{eq:12}, where $\mathbf{Q}_t^h$ represents the quality map inside the hand area of $\mathbf{x}_t''$.

\begin{equation} \label{eq:12}
\mathcal{L}_{pqgd} = -\mathbb{E}(\mathbf{Q}^h_{t-1})
\end{equation}

Finally, we leverage $\mathcal{L}_{pqgd}$ and the hand mask $\mathcal{M}_h$ to rapidly optimize the AQP within the hand region of $\mathbf{x}_t''$, as shown in Eq. \ref{eq:13}. Here, $\text{sgn}$ represents the sign function, which is used to compute the direction of the derivative of $\mathcal{L}_{pqgd}$ with respect to $\mathbf{x}''_{t-1}$, thereby accelerating optimization. The parameter $\delta_{pqgd}$ represents the learning rate of PQGD. The parameter $\epsilon$, similar to $\epsilon$ in PGD \cite{c22}, denotes the projection restriction parameter of PQGD, which constrains $\mathbf{x}_t''$ from deviating excessively from $\mathbf{x}'$ during optimization. This ensures that the additional PQGD perturbation only slightly alters the pixel values of AQP (such that the modification remains nearly imperceptible to the human eye), thereby preserving the effectiveness of the original AQP. It is important to emphasize that the optimization process is guided by $\mathcal{M}_h$ to operate solely within the hand region, endowing AQP with the adaptability to the human hand shape, which constitutes the most critical aspect of PQGD optimization.

\begin{equation} \label{eq:13}
\mathbf{x}_t'' = \left\{\prod_{\mathbf{x}', \epsilon}[\mathbf{x}_{t-1}''- \text{sgn}(\delta_{pqgd}\frac{\partial\mathcal{L}_{pqgd}}{\partial\mathbf{x}_{t-1}''})]\right\}\mathcal{M}_h + \mathbf{x}'(1-\mathcal{M}_h)
\end{equation}

\subsection{Active Adversarial for Robot Grasping} \label{sec:AA}

This part explains how QFAAP is applied to robot grasping to manipulate the quality score, enabling the robot to avoid grasping human hands and nearby objects. In our previous work \cite{c51}, we observed an intriguing property and empirically confirmed it through extensive real experiments that moving a specific object in a cluttered scenario can dynamically alter the quality score of this scenario. Specifically, if this object has a higher quality score, it can perturb objects with lower quality scores when the distance between them is very close (approximately 0.5–1 cm), leading to a further reduction in their quality scores. Moreover, as this object with the high quality score approaches, the quality scores of the affected objects will gradually decrease, and when they come into contact, the quality scores of these objects may drop sharply to zero. Notably, this phenomenon only occurs between adjacent objects; if the objects are far apart, no interference will happen, and their quality scores will remain unchanged. During the HRI process, the human hand can be regarded as a dynamically moving object. Thus, we are motivated to explore whether this property can be leveraged to enhance grasping safety in cluttered HRI scenarios.

QFAAP follows the property observed by \cite{c51}, processing the features within the human hand to increase its quality score using AQP and PQGD. Consequently, the human hand can be directly regarded as a benign adversarial perturbation that is actively against adjacent objects in any posture, thereby suppressing their quality scores. After the interference, the quality score within the human hand will be set to zero, reducing the grasping priority of both the hand and its adjacent objects. In other words, the manipulation of the quality score by QFAAP is entirely controllable and does not affect the original performance of the grasping model.

First, we use $\mathcal{M}_h$ to process $\mathbf{Q}_t$ from Section \ref{sec:PQGD}, setting the quality score within the hand region to zero. This results in a quality map outside the hand area of $\mathbf{x}_t''$, denoted as $\tilde{\mathbf{Q}}_t^h$. The robot then uses the perturbed $\tilde{\mathbf{Q}}_t^h$ as a reference and selects the object (away from the human hand and its adjacent objects) corresponding to the highest quality score in $\tilde{\mathbf{Q}}_t^h$ as the optimal grasping target. This process is defined in Eq.~\ref{eq:14}. Here, $(i_t^*, j_t^*)$ corresponds to the previously defined grasp candidate position parameters $(j^g, k^g)$, with the distinction that $(i_t^*, j_t^*)$ represents the optimal grasping position after QFAAP perturbation (where $t$ is to emphasize the influence of QFAAP). Furthermore, based on $(i_t^*, j_t^*)$, other optimal grasping parameters $w_t^*$, $h_t^*$, and $\theta_t^*$ can be determined, forming the optimal grasp $g_t^*$.

\begin{equation} \label{eq:14}
(i_t^*, j_t^*) = \underset{\substack{(i_t, j_t) \in (H, W)}}{\arg\max \tilde{\mathbf{Q}}_t^h(i_t, j_t)}
\end{equation}

Next, $g_t^*$ needs to undergo the following transformations to complete the grasping. Since $h_t^*$ is used only for visual representation and not in the conversion process, we denote the transferred optimal grasp in the robot end effector coordinate systems as ${\mathcal{G}_t^*} ({I}_t^*, {J}_t^*, {Z}_t^*, {W}_t^*, \Theta_t^*)$, which corresponds to the previously defined ${\mathcal{G}_i}(I^g, J^g, Z^g, W^g, \Theta^g)$, $t$ and $*$ are intended to emphasize the impact of QFAAP and optimal grasp. Here, $({I}_t^*, {J}_t^*, {Z}_t^*)$ represents the grasp position in the robot end effector coordinate system, ${W}_t^*$ is the opening stroke of the parallel jaw gripper, and $\Theta_t^*$ is the rotation angle of the gripper relative to the $Z$ axis. The conversion process is divided into two parts. The first part involves converting $(i_t^*, j_t^*)$: using depth information ($d$) and the camera's intrinsic parameters ($f_x$, $f_y$ for focal lengths and $c_x$, $c_y$ for the image center coordinates), we convert $(i_t^*, j_t^*)$ from the image coordinate system to the camera coordinate system $(i_{ct}^*, j_{ct}^*, z_{ct}^*)$, as shown in Eq. \ref{eq:15}. 

\begin{algorithm}[!b]
\caption{Quality-focused Active Adversarial Policy}
\label{alg:qfaap}
\begin{algorithmic}[1]
    \STATE \textbf{Input:}  Training sample $x_i$, real-time RGB frame $\mathbf{x}$ 
    \STATE \textbf{Output:} Optimal grasp in the robot end effector coordinate system $\mathcal{G}_t^*$\\
    // Adversarial Quality Patch: Using sample $x_i$ from grasp dataset $\mathbb{D}$, and solve Eq. \ref{eq:9} to optimize AQP.\\
    \FOR{$x_i \in \mathbb{D}$}
        \STATE $\mathbf{p}_t \leftarrow \mathcal{L}_{aqp}, \delta_{aqp}, x_i$
    \ENDFOR\\
    // Projected Quality Gradient Descent : First, $\mathbf{p}_t$ is added to the hand region by $\mathcal{M}_h$, generating $\mathbf{x}'$. Then, shape-adaptive optimization of AQP is performed by solving Eq.~\ref{eq:13}, yielding $\mathbf{x}''$. Finally, $\mathbf{x}_t''$ is fed into the grasping model to obtain $\mathbf{Q}_t$, along with the quality map $\tilde{\mathbf{Q}}_t^h$ outside the hand region after guided by $\mathcal{M}_h$.\\
    \STATE $\mathbf{x}' \leftarrow \mathbf{x}, \mathbf{p}_t, \mathcal{M}_h$
    \STATE $\mathbf{x}'' \leftarrow \mathbf{x}'_{t-1}, \mathbf{x}', \mathcal{L}_{pqgd}, \delta_{pqgd}, \mathcal{M}_h, \varepsilon$
    \STATE $\mathbf{Q}_t \leftarrow \mathbf{x}_t''$
    \STATE $\tilde{\mathbf{Q}}_t^h \leftarrow \mathbf{Q}_t, \mathcal{M}_h$\\
    // Active Adversarial for Robot Grasping: First, based on $\tilde{\mathbf{Q}}_t^h$, the grasp position $(i_t^*, j_t^*)$ corresponding to the maximum quality score is computed. Then, the remaining grasp parameters are obtained using $(i_t^*, j_t^*)$ to form the optimal grasp $g_t^*$. Finally, $g_t^*$ is transformed into the optimal grasp $\mathcal{G}_t^*$ in the robot end effector coordinate system by solve Eq. \ref{eq:15}, Eq. \ref{eq:16}, Eq. \ref{eq:17}.\\
    \FOR{$(i_t, j_t) \in (H, W), i_t \neq j_t$}
        \STATE $(i_t^*, j_t^*) \leftarrow{\arg\max \tilde{\mathbf{Q}}_t^h(i_t, j_t)}$
    \ENDFOR
    \STATE $g_t^* \leftarrow (i_t^*, j_t^*), w_t^*, h_t^*, \theta_t^*$
    \STATE $\mathcal{G}_t^* \leftarrow g_t^*$
    \RETURN $\mathcal{G}_t^*$
\end{algorithmic}
\end{algorithm}

\begin{equation} \label{eq:15}
\begin{bmatrix}
i_{ct}^* \\
j_{ct}^* \\
z_{ct}^*
\end{bmatrix} = 
\begin{bmatrix}
f_x^{-1} & 0 & -c_x f_x^{-1} \\
0 & f_y^{-1} & -c_y f_y^{-1} \\
0 & 0 & 1
\end{bmatrix} 
\begin{bmatrix}
i_{t}^* \\
j_{t}^* \\
1
\end{bmatrix}  {d}
\end{equation}

This is followed by converting $(i_{ct}^*, j_{ct}^*, z_{ct}^*)$ to the robot end effector coordinate system $({I}_t^*, {J}_t^*, {Z}_t^*)$ using the positional transformation relationship $\mathcal{T}$, as shown in Eq. \ref{eq:16}.

\begin{equation} \label{eq:16}
\begin{aligned}
({I}_t^*, {J}_t^*, {Z}_t^*) = \mathcal{T}(i_{ct}^*, j_{ct}^*, z_{ct}^*)
\end{aligned}
\end{equation}

The second part involves converting $(w_t^*, \theta_t^*)$ into $({W}_t^*, \Theta_t^*)$ using the projection function $\mathcal{P}$, as shown in Eq. \ref{eq:17}.

\begin{equation} \label{eq:17}
\begin{aligned}
({W}_t^*, \Theta_t^*) = \mathcal{P}({w}_t^*, \theta_t^*)
\end{aligned}
\end{equation}

The intrinsic parameters and depth information are directly obtained from the depth camera, and $\mathcal{T}$ is derived from offline eye-in-hand calibration. Finally, the projection function $\mathcal{P}$ allows for manual adjustment of the linear relationship between the gripper stroke $W_t^*$ and rotation $\Theta_t^*$ relative to the grasp box's width $w_t^*$ and rotation $\theta_t^*$.

Once the final grasp pose in the robot end effector coordinate system $({I}_t^*, {J}_t^*, {Z}_t^*, \Theta_t^*, \Theta_{xt}^*, \Theta_{yt}^*)$  is obtained, where $\Theta_{xt}^*$ and $\Theta_{yt}^*$ represent the constant rotations relative to the $X$-axis and the $Y$-axis, the gripper is moved to the target pose using inverse kinematics and its stroke is kept to the width ${W}_t^*$, thus achieving the avoidance of human hands and adjacent objects without emergency stops. The pseudocode of QFAAP is shown in Algorithm \ref{alg:qfaap}.

\section{Experiments}
In this section, we validate the effectiveness of our proposed method through extensive experiments. Firstly, we test the performance of AQP optimized by different grasping models and benchmark datasets. Then, we add PQGD to AQP to analyze the effectiveness of PQGD, as well as to explore the impact of iteration number on PQGD. Finally, we verify the performance of OFAAP on real robot grasping across different single-object and cluttered HRI scenarios.

\subsection{Experimental Settings}
\subsubsection{Setting for QFAAP} \label{sec:QFAAP}
We employ the Cornell Grasp Dataset \cite{c30}, Jacquard Grasp dataset \cite{c52}, and OCID Grasp Dataset \cite{c53}. The Cornell Grasp Dataset and Jacquard Grasp datasets are single-object RGB-D datasets, while OCID are cluttered RGB-D datasets. Cornell comprises 885 RGB-D images with a resolution of 640$\times$480, 240 different real objects, and 5k annotations. Jacquard is bigger than Cornell, with over 11k distinct simulated objects, 4900k annotations,  and 50k RGB-D images (1024$\times$1024). OCID \cite{c54}, designed to evaluate semantic segmentation methods in complex scenarios, provides diverse settings, including objects, backgrounds, lighting conditions, and so on. Therefore, we utilized an improved version from \cite{c53} for the grasping model, consisting of over 1.7k RGB-D images (640$\times$480) and 75k annotations. 

We train these DNNs-based grasping models in advance, thus can leveraging them for the optimization of AQP: GG-CNN~\cite{c31}, GG-CNN2 \cite{c32}, GR-ConvNet \cite{c34}, FCG-Net \cite{c35}, SE-ResUNet \cite{c33}, and TF-Grasp \cite{c55}. GR-ConvNet, FCG-Net, SE-ResUNet, and TF-Grasp support RGB images as input, while GG-CNN and GG-CNN2 accept Depth information. In our experiments, we extend GG-CNN and GG-CNN2 to handle RGB inputs by adjusting the number of input channels. These models were trained on a single NVIDIA RTX 4090 GPU with 24 GB of memory. The computer system is Ubuntu 22.04, and the deep learning framework is PyTorch 2.3.1 with CUDA 12.1. We follow the same image-wise setting in GR-ConvNet \cite{c34}, randomly shuffling the entire dataset, selecting 90\% for training and 10\% for testing before training. During training stage, the data will be uniformly cropped to 224$\times$224 (GG-CNN and GG-CNN2 are 300$\times$300), the total number of epochs for training is set to 50, the learning rate $\delta_{model}$ is fixed to 0.001, batch size $B$ is set to 8, and data augmentation (random zoom and random rotation) is applied (except Jacquard Grasp dataset). Finally, we employ the same rectangle (box) metric from \cite{c48} to assess the model performance, denoted as Original Accuracy (O-Acc). According to this metric, a predicted grasp by the grasping model is considered valid when it satisfies two conditions: the Intersection over Union score between the ground truth and predicted grasp rectangles is over 25\%, and the offset between the orientation of the ground truth rectangle and that of the predicted grasp rectangle is less than \SI{30}{\degree}.

For the optimization of AQP, we use the same device, system, and training parameters as the grasping model. Differently, we first initialize an AQP with a uniform distribution of size 224$\times$224 (300$\times$300 for GG-CNN and GG-CNN2). Next, during each iteration, we apply a random scale (ranging from 0.1 to 1 of the original size) to the AQP and paste it onto a random position of the training sample. We set $\alpha$, $\beta$, and $\gamma$ in $\mathcal{L}_q^p$ and $\mathcal{L}_{aqp}$ to 0.1, 0.1, and 0.5, respectively. The initial learning rate $\delta_{aqp}$ is set to 0.03 (decreasing by a factor of ten at the 30th and 40th epochs). It is important to note that since AQP does not need to be printed in the real world, as required by adversarial patch attacks, no additional data augmentation operations for AQP are used. Finally, we evaluate the performance of the AQP on the test set using the previously defined Q-ACC.

For the operation of PQGD, since it only processes real-time RGB frames, we only need to set the following parameters: the iteration number $N^i$ is set to 1, the learning rate $\delta_{pqgd}$ is fixed at 0.008, and $\epsilon$ is set to 8/255. In addition, we use the pre-trained model from \cite{c50} for real-time hand segmentation to guide the PQGD optimization. Finally, since PQGD is based on AQP, we use the same Q-ACC to evaluate the performance of PQGD.

\begin{figure}[!t]
\vspace{0.3\baselineskip}
\centerline{\includegraphics[width=\columnwidth]{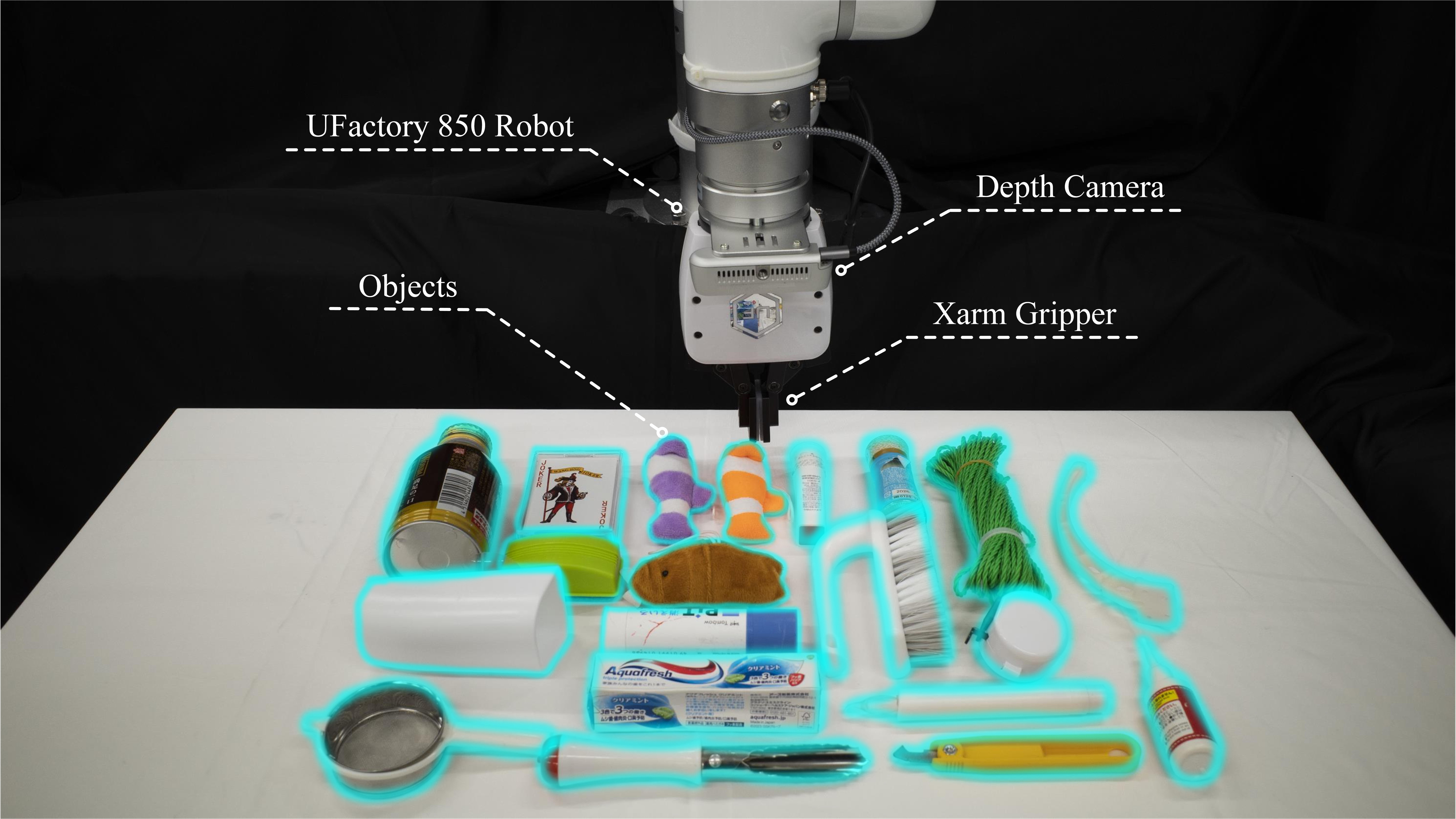}}
\caption{Experimental setup of robot grasping: primarily consisting of an Intel RealSense D435 depth camera, a UFactory 850 robot, a UFactory xArm gripper, and 20 novel objects (emphasized by blue borders).}
\label{fig3}
\end{figure}

\begin{table}[!t]
\caption{Results of AQP on the Cornell grasp dataset\label{tab:table1}}
\centering
\begin{tabular}{cccc}
\toprule 
Methods & O-ACC (\%) & Q-ACC (\%) & Speed ($s$)\\
\midrule
GG-CNN & 87.6 & 99.4 & 0.003\\
GG-CNN2 & 92.1 & 71.4 & 0.003\\
GR-Convnet & 96.6 & 94.2 & 0.005\\
FCG-Net & 96.6 & 97.4 & 0.009\\
SE-ResUNet & 95.5 & 90.4 & 0.013\\
TF-Grasp & 96.8 & 27.0 & 0.008\\
\bottomrule
\end{tabular}
\end{table}

\begin{table}[!t]
\caption{Results of AQP on the OCID grasp dataset\label{tab:table2}}
\centering
\begin{tabular}{cccc}
\toprule 
Methods & O-ACC (\%) & Q-ACC (\%) & Speed ($s$)\\
\midrule
GG-CNN & 18.6 & 96.9 & 0.003\\
GG-CNN2 & 44.6 & 90.0 & 0.003\\
GR-Convnet & 53.7 & 93.9 & 0.006\\
FCG-Net & 52.5 & 91.1 & 0.008\\
SE-ResUNet & 46.3 & 98.5 & 0.014\\
TF-Grasp & 26.0 & 94.1 & 0.007\\
\bottomrule
\end{tabular}
\end{table}

\begin{table}[!t]
\caption{Results of AQP on the Jacquard grasp dataset\label{tab:table3}}
\centering
\begin{tabular}{cccc}
\toprule 
Methods & O-ACC (\%) & Q-ACC (\%) & Speed ($s$)\\
\midrule
GG-CNN & 83.7 & 74.8 & 0.004\\
GG-CNN2 & 86.0 & 71.5 & 0.004\\
GR-Convnet & 91.8 & 70.9 & 0.007\\
FCG-Net & 86.3 & 79.3 & 0.011\\
SE-ResUNet & 85.5 & 82.3 & 0.017\\
TF-Grasp & 93.6 & 51.3 & 0.013\\
\bottomrule
\end{tabular}
\end{table}

\subsubsection{Setting for Robot Grasping} \label{sec:Egrasp}
Our robot grasping system is illustrated in Fig. \ref{fig3}. In particular, we adopt an eye-in-hand grasping architecture, where the camera is fixed on the robot, and the field of view faces downward. In addition, we define the following evaluation criteria to assess the effectiveness of our method in the real world, including the success rate of detecting optimal grasps that do not occur on the hand or its adjacent objects (ND-ACC) and the collision rate of the robot to the hand during the grasping process(CH-Rate). It is important to emphasize that due to the presence of the hand in various scenarios, this grasping experiment may cause human injury. Therefore, we fix the robot at a safe height (other predicted position parameters by the grasping model remain unchanged) and then slowly move the robot to the actual height during each grasping.

\subsection{Effectiveness of AQP} \label{sec:EAQP}
We employ the same experimental setting of AQP and grasping model discussed in Section \ref{sec:QFAAP}, with the corresponding results presented in Table \ref{tab:table1} (optimized using the Cornell Grasp dataset), Table \ref{tab:table2} (optimized using the OCID Grasp dataset), and Table \ref{tab:table3} (optimized using the Jacquard Grasp dataset). To ensure consistency and avoid confusion, we refer to some results reported in the original papers, such as the O-Acc of GR-ConvNet \cite{c34} and TF-Grasp \cite{c55} trained on the Cornell and Jacquard Grasp datasets. In Table \ref{tab:table1}, AQP  optimized by most models can achieve a Q-AAC exceeding 90\%, except for optimized by GG-CNN2, which attains 71.4\%, and TF-Grasp, which records 27.0\%. In Table \ref{tab:table2}, AQP optimized by all models exhibits a Q-AAC above 90\%. In Table \ref{tab:table3}, despite being optimized using a large-scale dataset (with extensive test images for testing), AQP optimized by most models can still surpass 70\%, except for optimized by TF-Grasp, which gets 51.3\%.

The above analyses indicate that AQP optimized across different datasets and models is effective. Furthermore, AQP optimized using cluttered datasets demonstrates superior performance compared to single-object datasets, providing a solid foundation for the subsequent application of QFAAP in cluttered grasping scenarios. Finally, we visualize the quality performance of AQP across these datasets in the first two rows of Fig. \ref{fig5}, Fig. \ref{fig6}, and Fig. \ref{fig7}. As illustrated in this figure, although the highest quality scores are not located on AQP in columns 3 and 5-8 of Fig. \ref{fig6}, as well as columns 1, 2, and 5 of Fig. \ref{fig7}, most highest scores are concentrated on AQP, further demonstrating the effectiveness of AQP in manipulating the quality score.
 
\subsection{Effectiveness of PQGD}

We validate PQGD by applying it to the AQP optimized in Section \ref{sec:EAQP} and employing the experimental settings of PQGD discussed in Section \ref{sec:QFAAP}. In addition, the iteration number $N^i$ is set to 1 in this part. The experimental results are presented in Table \ref{tab:table4}, Table \ref{tab:table5}, and Table \ref{tab:table6}. By comparing these tables with their corresponding Table \ref{tab:table1}, Table \ref{tab:table2}, and Table \ref{tab:table3}, it can be observed that PQGD consistently improves the quality score of the AQP optimized by all models and datasets, with a more pronounced effect on the Jacquard Grasp dataset, resulting in an overall quality score improvement of approximately 2\%. Although the prediction speed decreases with adding PQGD, it remains close to real-time performance. This reduction has no impact on the efficiency of robot grasping, as the movement time of the robot is significantly longer than the prediction time of the grasping model in practice. Therefore, we enable AQP to rapidly acquire the human hand shape adaptability at a low cost. 

\begin{figure}[!t]
\vspace{0.3\baselineskip}
\centerline{\includegraphics[width=\columnwidth]{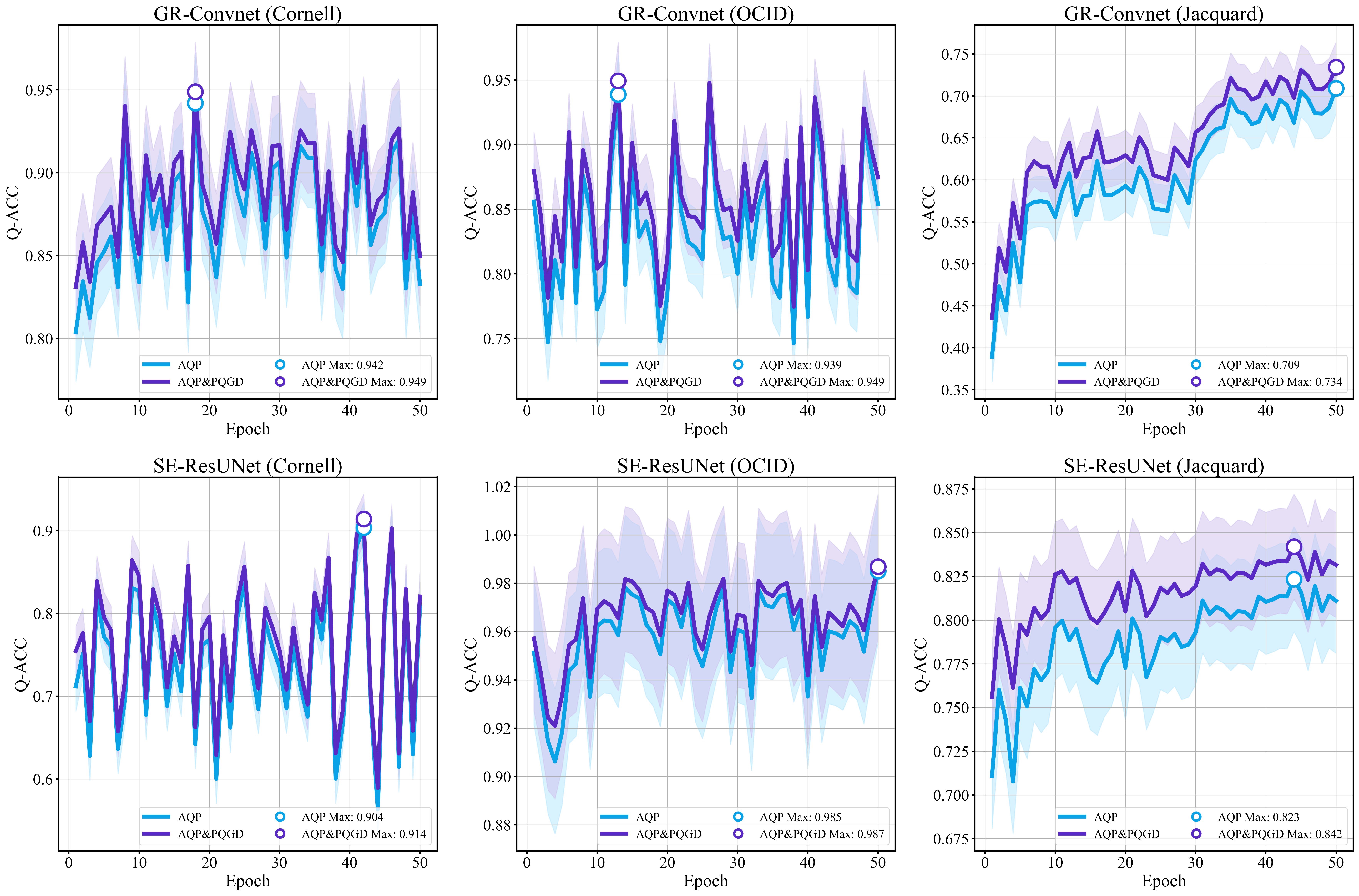}}
\caption{Line graphs showing the effectiveness of PQGD across all epochs, including its impact on the AQP optimized by GR-ConvNet and three different datasets, as well as the AQP optimized by SE-ResUNet and three different datasets. Here, the AQP and AQP\&PQGD are represented by blue and purple lines, and we also use blue and purple dots to emphasize their corresponding maximum quality score across all epochs.}
\label{fig4}
\end{figure}

\begin{figure*}[!t]
\centerline{\includegraphics[width=\textwidth]{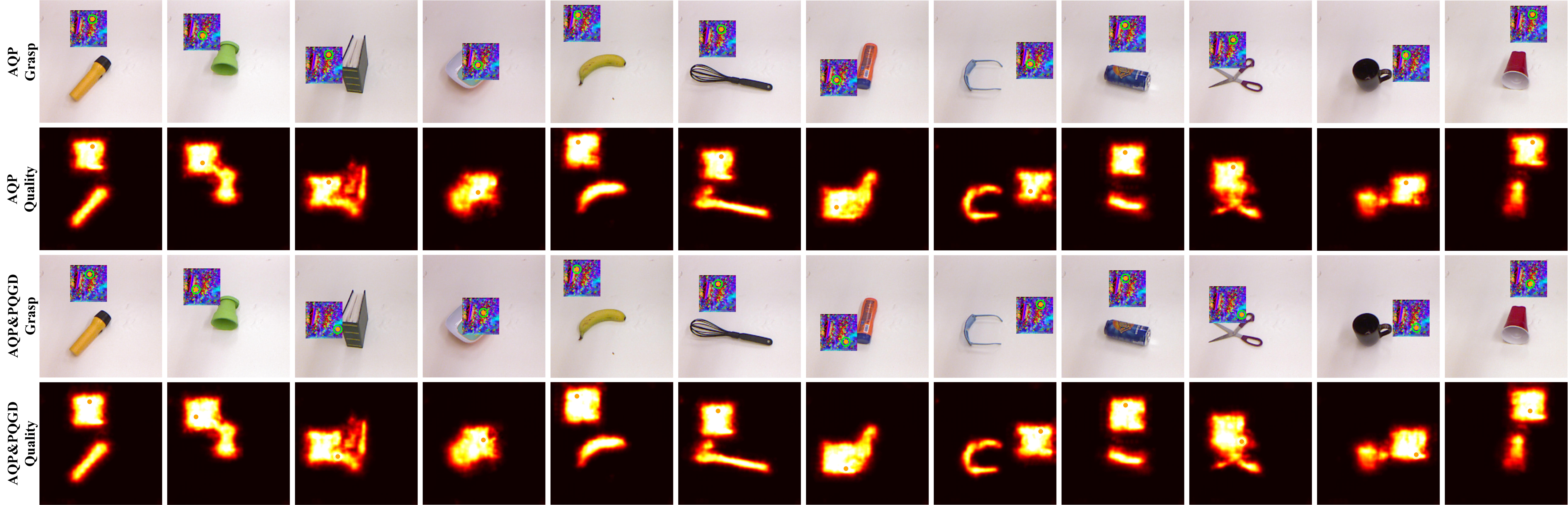}}
\caption{Quality score visualization of AQP (first two rows) before and after adding PQGD (last two rows). Here, the GGCNN2 and the Cornell Grasp dataset are used to optimize the AQP. The AQP before and after adding PQGD are located in different locations. And AQP is scaled to 0.3 of the original size (the same size of the image).}
\label{fig5}
\end{figure*}

\begin{figure*}[!t]
\centerline{\includegraphics[width=\textwidth]{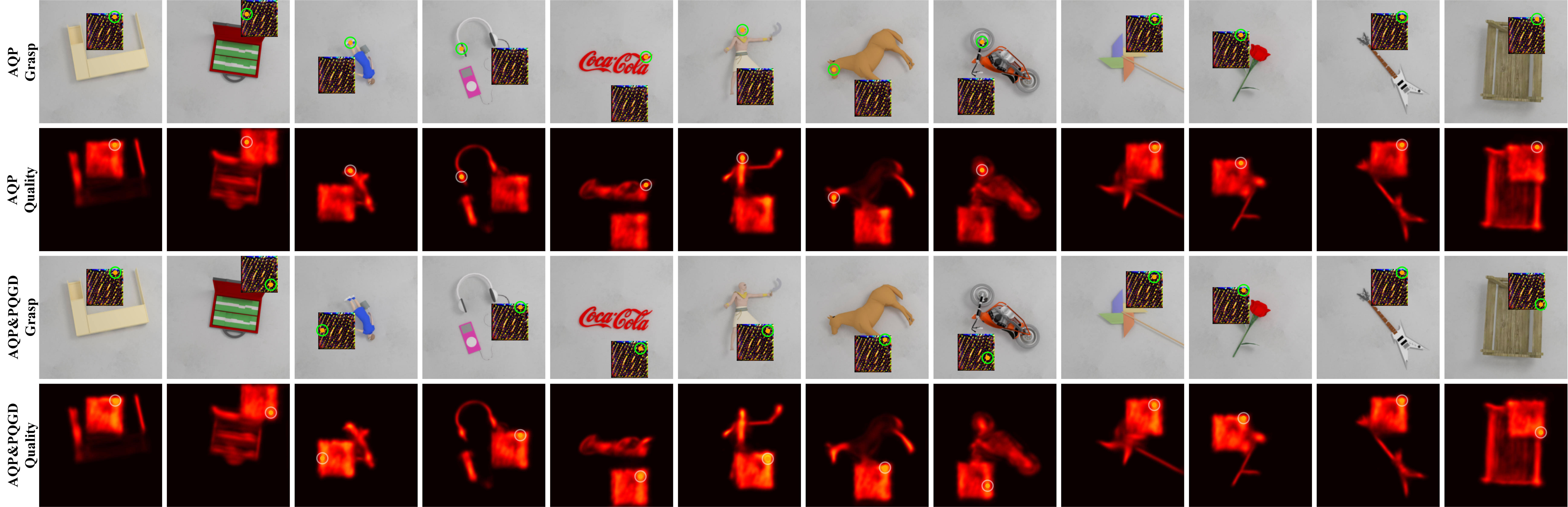}}
\caption{The meaning of each row is consistent with Fig. \ref{fig5}. Here, the SE-ResUNet and the Jacquard Grasp dataset are used to optimize the AQP. The AQP before and after adding PQGD are located in different locations. And AQP is scaled to 0.3 of the original size (the same size of the image).}
\label{fig6}
\end{figure*}

\begin{figure*}[!t]
\centerline{\includegraphics[width=\textwidth]{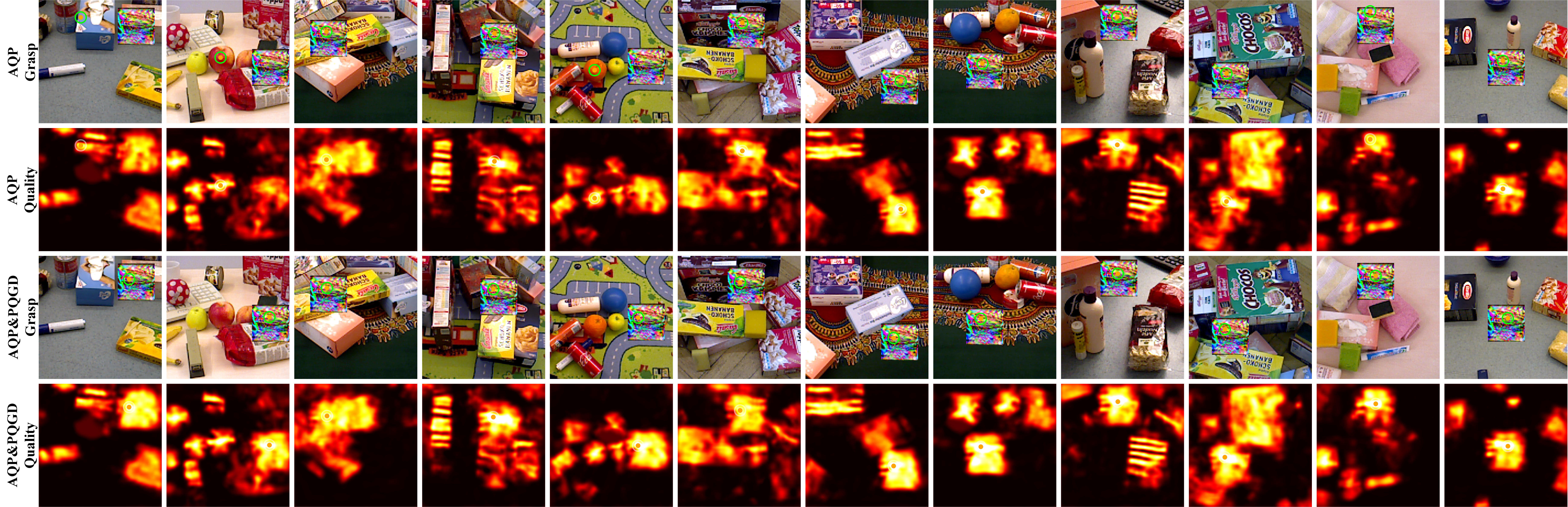}}
\caption{The meaning of each row is consistent with Figs. \ref{fig5} and \ref{fig6}. Here, the GR-ConvNet and the OCID Grasp dataset are used to optimize the AQP. The AQP before and after adding PQGD are located in different locations. The size is 0.3 times the original size. And AQP is scaled to 0.3 of the original size (the same size of the image).}
\label{fig7}
\end{figure*}

Additionally, we show the effectiveness of PQGD across all epochs in Fig.~\ref{fig4}. As illustrated in this figure, it is evident that PQGD remains effective throughout all epochs. Since we applied only a random scale to AQP without additional augmentations, the quality score exhibits fluctuations on the smaller Cornell Grasp and OCID Grasp datasets due to overfitting. However, this issue is eliminated for the larger Jacquard Grasp dataset. Overall, this fluctuation does not impact the subsequent deployment of our QFAAP, as our objective is not to attack the model but to ensure the achievement of a high quality score. We also visualize the quality performance of AQP after adding PQGD across these datasets in the last two rows of Figs. \ref{fig5}, \ref{fig6}, and \ref{fig7}. As shown in this figure, PQGD can further improve the manipulation of the quality score by AQP.

\begin{table}[!t]
\caption{Results of AQP\&PQGD on the Cornell grasp dataset\label{tab:table4}}
\centering
\begin{tabular}{cccc}
\toprule 
Methods & O-ACC (\%) & Q-ACC (\%) & Speed ($s$)\\
\midrule
GG-CNN & 87.6 & 99.5 & 0.011\\
GG-CNN2 & 92.1 & 72.3 & 0.016\\
GR-Convnet & 96.6 & 94.9 & 0.031\\
FCG-Net & 96.6 & 97.6 & 0.042\\
SE-ResUNet & 95.5 & 91.4 & 0.056\\
TF-Grasp & 96.8 & 31.3 & 0.038\\
\bottomrule
\end{tabular}
\end{table}

\begin{table}[!t]
\caption{Results of AQP\&PQGD on the OCID grasp dataset\label{tab:table5}}
\centering
\begin{tabular}{cccc}
\toprule 
Methods & O-ACC (\%) & Q-ACC (\%) & Speed ($s$)\\
\midrule
GG-CNN & 18.6 & 97.6 & 0.012\\
GG-CNN2 & 44.6 & 93.0 & 0.017\\
GR-Convnet & 53.7 & 94.9 & 0.031\\
FCG-Net & 52.5 & 92.4 & 0.044\\
SE-ResUNet & 46.3 & 98.7 & 0.058\\
TF-Grasp & 26.0 & 94.7 & 0.033\\
\bottomrule
\end{tabular}
\end{table}

\begin{table}[!t]
\caption{Results of AQP\&PQGD  on the Jacquard grasp dataset\label{tab:table6}}
\centering
\begin{tabular}{cccc}
\toprule 
Methods & O-ACC (\%) & Q-ACC (\%) & Speed ($s$)\\
\midrule
GG-CNN & 83.7 & 76.0 & 0.017\\
GG-CNN2 & 86.0 & 74.6 & 0.023\\
GR-Convnet & 91.8 & 73.4 & 0.037\\
FCG-Net & 86.3 & 82.2 & 0.052\\
SE-ResUNet & 85.5 & 84.2 & 0.069\\
TF-Grasp & 93.6 & 57.1 & 0.069\\
\bottomrule
\end{tabular}
\end{table}

\subsection{Impact of Iteration Number on PQGD}

\begin{figure*}[!t]
\centerline{\includegraphics[width=\textwidth]{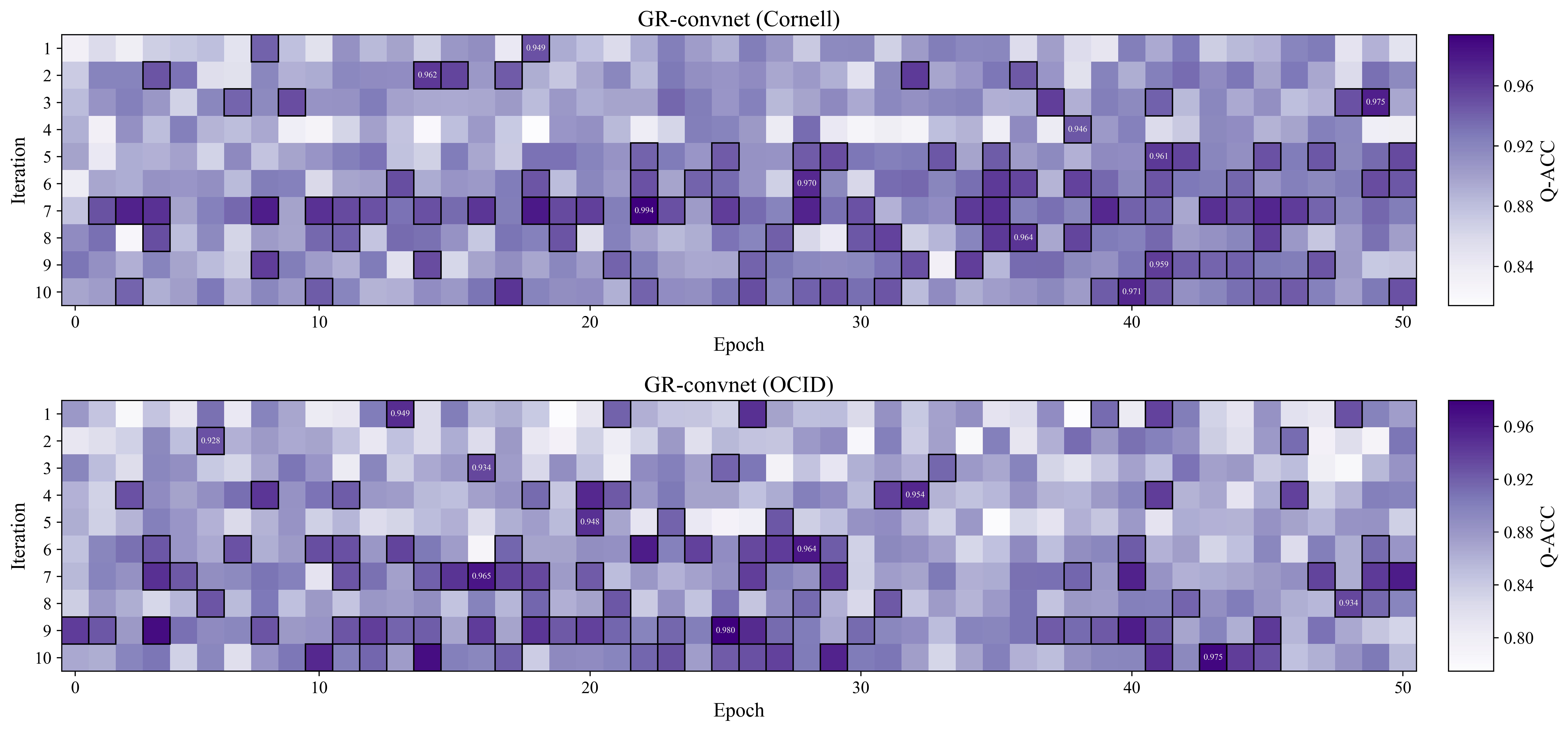}}
\caption{Heatmap showing the impact of the iteration number $N^i$ on PQGD across all epochs. Here, the AQP is optimized by GR-ConvNet on the Cornell Grasp dataset (upper sub-figure) and the OCID Grasp dataset (lower sub-figure). In addition, the maximum quality score for each row is printed in white numbers for emphasis.}
\label{fig8}
\end{figure*}

This part primarily investigates the impact of the iteration number $N^i$ on PQGD. We conduct experiments using the AQP optimized by GR-ConvNet on the Cornell Grasp dataset and the OCID Grasp dataset, with the iteration number $N^i$ ranging from 1 to 10. Other experimental settings remain the same as in Section \ref{sec:QFAAP}. The results are presented in Table \ref{tab:table7}, which shows that the optimal number of iterations for PQGD is around 7 for the Cornell Grasp dataset and around 9 for the OCID Grasp dataset. Overall, different numbers of iterations consistently lead to an improvement in Q-ACC. Additionally, we visualize the effect of the number of iterations $N^i$ on PQGD across all epochs in Fig. \ref{fig8}, and the conclusions drawn from the visualization are consistent with the above statements.

\subsection{Effectiveness of QFAAP in Robot Grasping}\label{sec:robot}

\begin{table*}[!t]
\caption{The impact of different iteration numbers of PQGD on Q-ACC\label{tab:table7}}
\centering
\begin{tabular}{cccccccccccc}
\toprule 
Iteration Number $N^i$  & 1 & 2 & 3 & 4 & 5 & 6 & 7 & 8 & 9 & 10\\
\midrule
Cornell Q-ACC (\%) & 94.9 & 96.2 & 97.5 & 94.6 & 96.1 & 97.0 & 99.4 & 96.4 & 95.9 & 97.1\\
OCID Q-ACC (\%) & 94.9 & 92.8 & 93.4 & 95.4 & 94.8 & 96.4 & 96.5 & 93.4 & 98.0 & 97.5\\
\bottomrule
\end{tabular}
\end{table*}

\begin{figure*}[!t]
\vspace{0.3\baselineskip}
\centerline{\includegraphics[width=\textwidth]{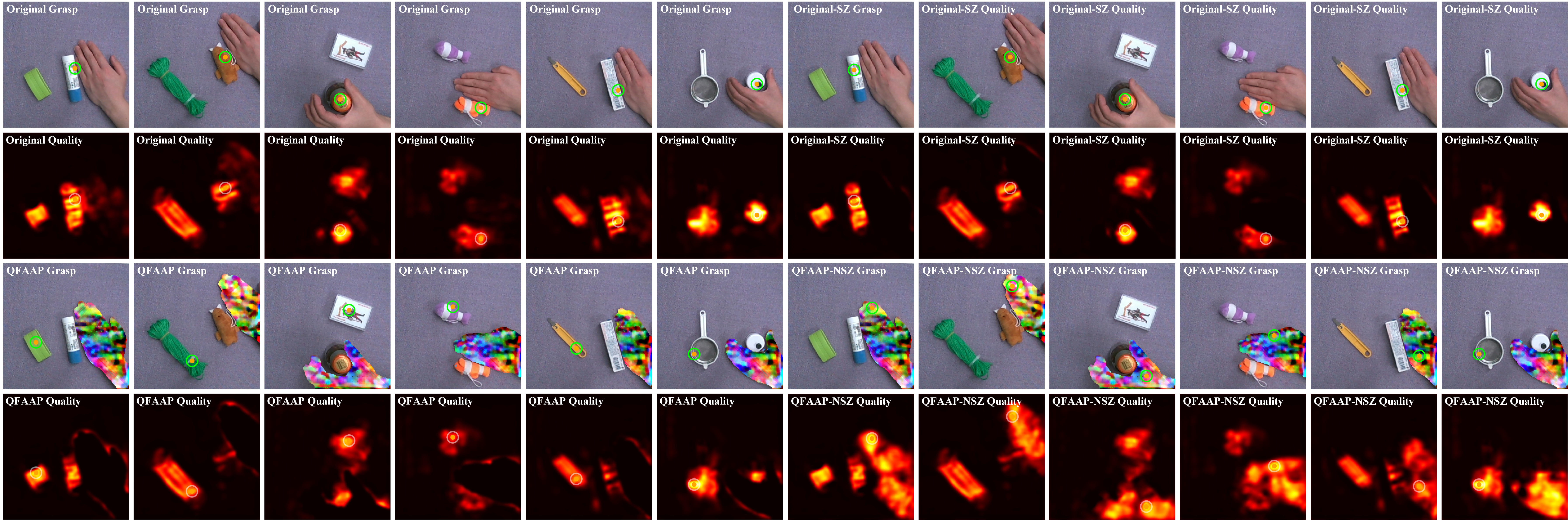}}
\caption{Visualization of optimal grasp and quality map for Original (first two rows of the first to sixth columns), Original-SZ (first two rows of the seventh to twelfth columns), QFAAP (last two rows of the first to sixth columns), and QFAAP-NSZ (last two rows of the seventh to twelfth columns).}
\label{fig9}
\end{figure*}

\begin{figure*}[!t]
\centerline{\includegraphics[width=\textwidth]{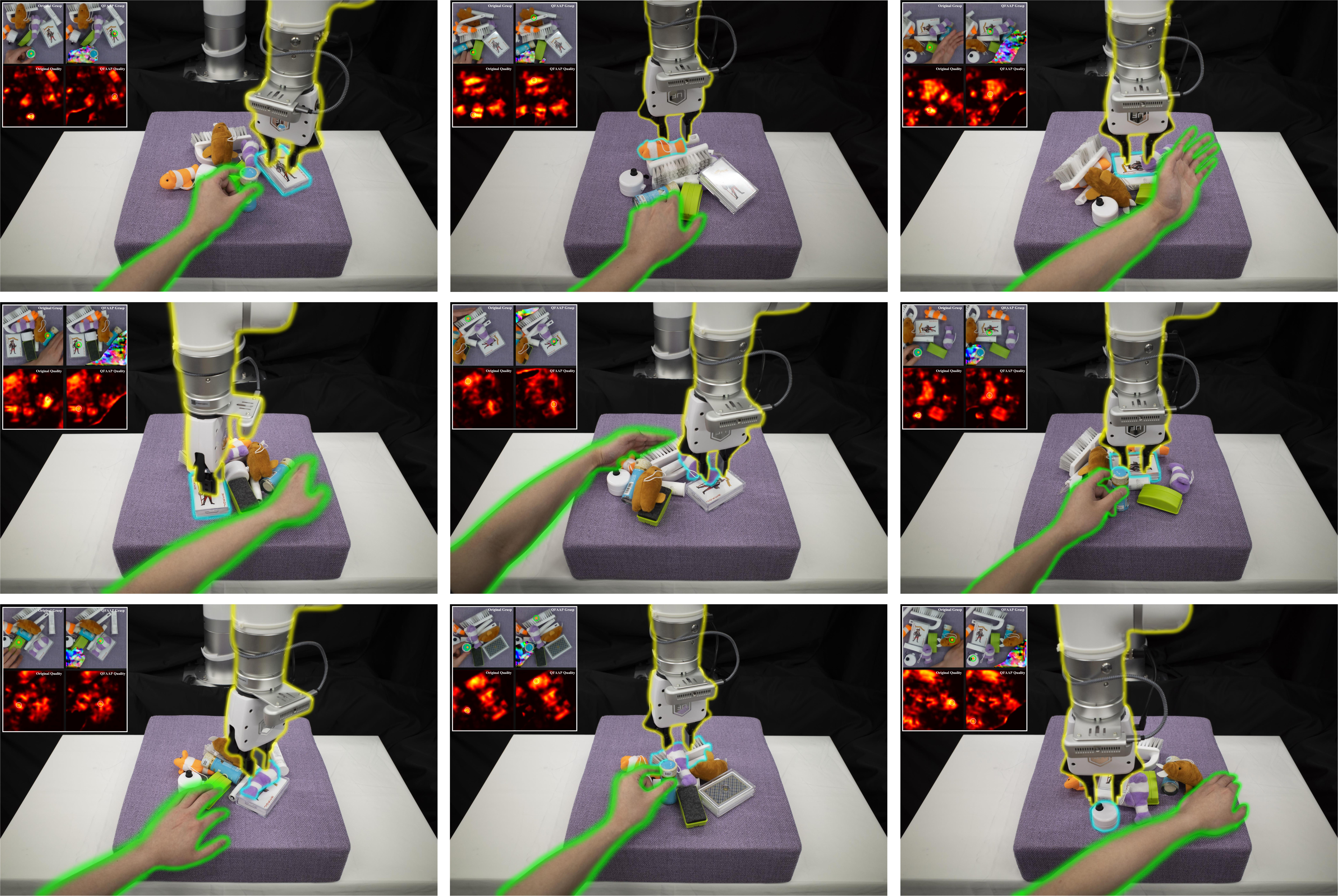}}
\caption{Grasping of QFAAP and the Original method. We use yellow, blue, and green borders to highlight the robot, the human hand, and the objects being grasped. In addition, we added the optimal grasp and quality map for QFAAP and the original method to each sub-figure.}
\label{fig10}
\end{figure*}

\subsubsection{Single Object Grasping Scenarios}\label{sec:single}

Here we evaluate the performance of QFAAP in real-world single-object scenarios. First, we group the experimental objects into ten pairs. To assess the effectiveness of QFAAP, we approach the object part with the highest quality score within each object pair ten times, where the object positions and human hand postures are randomly adjusted in each trial. The comparison methods include Original (the original grasping model), Original-SZ (a variant of the grasping model where the quality score of the hand region is set to zero), and our proposed method, QFAAP. In addition, we use the AQP optimized by GR-ConvNet and OCID Grasp dataset and set the iteration number $N^i$ to 5 for PQGD. All other experimental settings are consistent with Section \ref{sec:Egrasp}. The results are presented in Table \ref{tab:table8}. Our method significantly outperforms both Original and Original-SZ, achieving an ND-ACC of 88\%, demonstrating the shape adaptability of QFAAP and its effectiveness in enhancing the safety of the HRI process in single-object grasping scenarios. We also visualize some of our results in Fig. \ref{fig9}, including the optimal grasp and quality map for Original, Original-SZ, QFAAP, and QFAAP-NSZ (a variant of the QFAAP where the quality score of the hand region is not set to zero). The conclusion shown in this figure is consistent with the above statements. It should be noted that QFAAP-NSZ is only to emphasize the strength of the quality score for QFAAP and is not included in Table \ref{tab:table8}. Finally, the few failure cases of QFAAP primarily result from situations where the object approached by the human hand still maintains a higher quality score than the other object. Additionally, extremely rapid hand movements can lead to failures. In future work, we will enhance our optimization methods to strengthen QFAAP and incorporate human motion estimation to help QFAAP better adapt to dynamic changes in hand posture.

\begin{table*}[!t]
\caption{Detection results in single object scenarios\label{tab:table8}}
\centering
\begin{tabular}{ccccccccccccc}
\toprule 
Object Pairs & P1 & P2 & P3 & P4 & P5 & P6 & P7 & P8 & P9 & P10 & Overall (\%)\\
\midrule
Original ND-ACC & 1/10 & 0/10 & 2/10 & 0/10 & 1/10 & 1/10 & 1/10 & 1/10 & 3/10 & 3/10 & 13\\
Original-SZ ND-ACC & 1/10 & 0/10 & 3/10 & 0/10 & 1/10 & 2/10 & 1/10 & 1/10 & 3/10 & 3/10 & 15\\
QFAAP ND-ACC & 7/10 & 9/10 & 9/10 & 10/10 & 8/10 & 9/10 & 10/10 & 8/10 & 8/10 & 10/10 & 88\\
\bottomrule
\end{tabular}
\end{table*}

\begin{table*}[!t]
\caption{Grasping results in clutter scenarios\label{tab:table9}}
\centering
\begin{tabular}{ccccccccccccc}
\toprule 
Scenarios & S1 & S2 & S3 & S4 & S5 & S6 & S7 & S8 & S9 & S10 & Overall (\%)\\
\midrule
Original CH-Rate & 8/10 & 6/10 & 6/10 & 7/10 & 8/10 & 4/10 & 5/10 & 6/10 & 7/10 & 5/10 & 62\\
Original-SZ CH-Rate & 6/10 & 6/10 & 6/10 & 6/10 & 7/10 & 4/10 & 5/10 & 6/10 & 7/10 & 5/10 & 58\\
QFAAP CH-Rate & 2/10 & 1/10 & 2/10 & 3/10 & 2/10 & 0/10 & 2/10 & 1/10 & 2/10 & 1/10 & 16\\
\bottomrule
\end{tabular}
\end{table*}

\subsubsection{Clutter Grasping Scenarios}

We use a similar experimental setting as in Section \ref{sec:single} to evaluate the performance of QFAAP on a real robot grasping system in cluttered scenarios. Differently, we select 10 objects from the experimental objects to create 10 different cluttered grasping scenes. We perform 10 grasp attempts for each scene and adjust the human hand posture in each trial (with the object being approached with the highest quality score in the scene, too). Additionally, we primarily evaluate the collision rate between the robot and the human hand (CH-Rate).

The experimental results are shown in Table \ref{tab:table9}, where our method consistently outperforms both the Original and Original-SZ methods, achieving a notably low CH-Rate of 16\%. This result demonstrates the effectiveness of QFAAP in enhancing the safety of the HRI process in cluttered grasping scenarios. Furthermore, the reasons for the failure cases of QFAAP in this scenario remain consistent with those in Section \ref{sec:single}. Finally, we also visualize some results related to QFAAP and the Original method in Fig. \ref{fig10}, and the conclusion is consistent with the above statements.

\section{Conclusion}

In this paper, we proposed the Quality-focused Active Adversarial Policy (QFAAP), which first optimized an Adversarial Quality Patch (AQP) with high quality scores using the adversarial quality patch loss and a grasp dataset. Then, the Projected Quality Gradient Descent (PQGD) was introduced to optimize AQP further, endowing it with the adaptability to the human hand shape. By leveraging AQP and PQGD, the hand itself can be an active perturbation source against nearby objects, reducing their quality scores. Further setting the quality score of the hand to zero will reduce the grasping priority of both the hand and its adjacent objects, enabling the robot to avoid them without emergency stops for autonomous grasping. We conducted extensive experiments on the benchmark datasets and a cobot, showing that QFAAP can improve the safety of robot grasping both in single-object and cluttered HRI scenarios. 

Future work can be divided into two major parts. The first part can focus on addressing the issues highlighted in \ref{sec:robot} to enhance the method proposed in this paper. The second part involves exploring how to extend QFAAP to incorporate multimodal properties, which can then be utilized to address the backdoor attack problem proposed in \cite{c51}.


\phantomsection


\addcontentsline{toc}{section}{References} 

\begin{thebibliography}{99}
\bibitem{c1} R. M. Murray, Z. Li, and S. S. Sastry, \emph{A Mathematical Introduction to Robotic Manipulation}. Boca Raton, FL, USA: CRC Press, 2017.

\bibitem{c2} D. Prattichizzo and J. C. Trinkle, “Grasping,” in \emph{Springer Handbook of Robotics}, Berlin, Germany: Springer 2008.

\bibitem{c3} B. Kehoe, A. Matsukawa, S. Candido, J. Kuffner, and K. Goldberg, “Cloud-based robot grasping with the google object recognition engine,” in \emph{Proc. IEEE Int. Conf. Robot. Automat.}, 2013, pp. 4263–4270.

\bibitem{c4} J. Mahler et al., “Learning ambidextrous robot grasping policies,” \emph{Sci. Robot.}, vol. 4, no. 26, pp. 1–12, 2019.

\bibitem{c5} H. S. Fang, M. Gou, C. Wang, and C. Lu, “Robust grasping across diverse sensor qualities: The GraspNet-1Billion dataset,” \emph{Int. J. Robot. Res.}, vol. 42, no. 12, pp. 1094–1103, 2023.

\bibitem{c6} A. Vaswani, N. Shazeer, N. Parmar, J. Uszkoreit, L. Jones, A.N. Gomez, Ł. Kaiser, and I. Polosukhin, “Attention is all you need,” in \emph{Proc. Conf. Neural Informat. Process. Syst.}, 2017, pp. 6000–6010.

\bibitem{c7} S. Hochreiter, and J. Schmidhuber, "Long short-term memory," \emph{Neural Comput.}, vol. 9, no. 8, pp. 1735–1780, 1997.

\bibitem{c8} T. Brown, b. Mann, N. Ryder, M. Subbiah, J.D. Kaplan, P. Dhariwal, A. Neelakantan, P. Shyam, G. Sastry, A. Askell, and S. Agarwal, "Language models are few-shot learners," in \emph{Proc. Conf. Neural Informat. Process. Syst.}, 2020, pp. 1877–1901.

\bibitem{c9} A. Dosovitskiy, L. Beyer, A. Kolesnikov, D. Weissenborn, X. Zhai, T. Unterthiner, M. Dehghani, M. Minderer, G. Heigold, S. Gelly, and J. Uszkoreit, "An image is worth 16x16 words: Transformers for image recognition at scale," 2020, \emph{arXiv:}2010.11929.

\bibitem{c10} R. Girshick, J. Donahue, T. Darrell, and J. Malik, "Rich feature hierarchies for accurate object detection and semantic segmentation," in \emph{Proc. IEEE Conf. Comput. Vis. Pattern Recognit.}, 2014, pp. 580-587.

\bibitem{c11} C. Meng, T. Zhang, and T. l. Lam, “Fast and comfortable interactive robot-to-human object handover,” in \emph{Proc. IEEE/RSJ Int. Conf. Intell. Robots Syst.}, 2022, pp. 3701-3706.

\bibitem{c12} S. Christen, L. Feng, W. Yang, Y.-W. Chao, O. Hilliges, and J. Song, “SynH2R:Synthesizing hand-object motions for learning human-to-robot handovers,” in \emph{Proc. IEEE Int. Conf. Robot. Automat.}, 2024, pp. 3168–3175.

\bibitem{c13} H. Duan, P. Wang, Y. Yang, D. Li, W. Wei, Y. Luo, and G. Deng, “Reactive Human-to-Robot Dexterous Handovers for Anthropomorphic Hand,“ \emph{IEEE Trans. Robot.}, vol. 41, pp. 742 - 761, 2024.

\bibitem{c14} Z. Wang, J. Chen, Z. Chen, P. Xie, R. Chen, and L. Yi, “GenH2R: Learning generalizable human-to-robot handover via scalable simulation, demonstration, and imitation,” in \emph{Proc. IEEE Conf. Comput. Vis. Pattern Recognit.}, 2024, pp. 16362–16372.

\bibitem{c15} P. Rosenberger et al., “Object-independent human-to-robot handovers using real time robotic vision,” \emph{IEEE Robot. Automat. Lett.}, vol. 6, no. 1, pp. 17–23, 2021.

\bibitem{c16} I. J. Goodfellow, J. Shlens, and C. Szegedy, “Explaining and harnessing adversarial examples,” in \emph{Proc. Int. Conf. Learn. Representations.}, 2015.

\bibitem{c17} T. Long, Q. Gao, L. Xu, and Z. Zhou, “A survey on adversarial attacks in computer vision: Taxonomy, visualization and future directions,” \emph{Comput. Secur.}, vol. 121, pp. 102847, 2022.

\bibitem{c18} J. Wang et al., “PISA: Pixel skipping-based attentional black-box adversarial attack,” \emph{Comput. Secur.}, vol. 121, pp. 102947, 2022.

\bibitem{c19} Z. Wang, F. Nie, H. Wang, H. Huang, and F. Wang, “Toward robust discriminative projections learning against adversarial patch attacks,” \emph{IEEE Trans. Neural Netw. Learn. Syst.}, vol. 35, no. 12, pp. 18784 - 18798, 2024.

\bibitem{c20} G. Li, Y. Xu, J. Ding, and G.-S. Xia, “Towards generic and controllable attacks against object detection,” \emph{IEEE Trans. Geosci. Remote Sens.}, vol. 62, 2024.

\bibitem{c21} K.-H. Chow et al., “Adversarial objectness gradient attacks in real-time object detection systems,” in \emph{IEEE Int. Conf. Trust, Privacy Secur. Intell. Syst.}, 2020, pp. 263–272.

\bibitem{c22} A. Madry, A. Makelov, L. Schmidt, D. Tsipras, and A. Vladu, “Towards deep learning models resistant to adversarial attacks,” in \emph{Proc. Int. Conf. Learn. Representations.}, 2018.

\bibitem{c23} X. Cui, A. Aparcedo, Y. K. Jang, and S.-N. Lim, “On the robustness of large multimodal models against image adversarial attacks,” in \emph{Proc. IEEE Conf. Comput. Vis. Pattern Recognit.}, 2024, pp. 24625–24634.

\bibitem{c24} S. Thys, W. Van Ranst, and T. Goedeme, “Fooling automated surveillance cameras: Adversarial patches to attack person detection,” in \emph{Proc. IEEE Conf. Comput. Vis. Pattern Recognit. Workshops.}, 2019, pp. 49–55.

\bibitem{c25} Z. Hu, S. Huang, X. Zhu, F. Sun, B. Zhang, and X. Hu, “Adversarial texture for fooling person detectors in the physical world,” in \emph{Proc. IEEE Conf. Comput. Vis. Pattern Recognit.}, 2022, pp. 13307–13316.

\bibitem{c26} K. Xu et al., “Adversarial T-shirt! Evading person detectors in a physical world,” in \emph{Proc. Eur. Conf. Comput. Vis.}, 2020, pp. 665–681.

\bibitem{c27} Z. Hu, W. Chu, X. Zhu, H. Zhang, B. Zhang, and X. Hu, “Physically realizable natural-looking clothing textures evade person detectors via 3D modeling,” in \emph{Proc. IEEE Conf. Comput. Vis. Pattern Recognit.}, 2023, pp. 16975–16984.

\bibitem{c28} C. Rosales, J. M. Porta, and L. Ros, “Grasp optimization under specific contact constraints,” \emph{IEEE Trans. Robot.}, vol. 29, no. 3, pp. 746–757, 2013.

\bibitem{c29} F. T. Pokorny, K. Hang, and D. Kragic, “Grasp moduli spaces,” in \emph{Proc. Robot.: Sci. Syst.}, 2013.

\bibitem{c30} I. Lenz, H. Lee, and A. Saxena, “Deep learning for detecting robotic grasps,” \emph{Int. J. Robot. Res.}, vol. 34, no. 4–5, pp. 705–724, 2015.

\bibitem{c31} D. Morrison, P. Corke, and J. Leitner, “Closing the loop for robotic grasping: A real-time, generative grasp synthesis approach,” in \emph{Proc. Robot.: Sci. Syst.}, 2018.

\bibitem{c32} D. Morrison, P. Corke, and J. Leitner, “Learning robust, real-time, reactive robotic grasping,” \emph{Int. J. Robot. Res.}, vol. 39, no. 2-3, pp. 183–201, 2020.

\bibitem{c33} S. Yu, D.-H. Zhai, Y. Xia, H. Wu, and J. Liao, “SE-ResUNet: A novel robotic grasp detection method,” \emph{IEEE Robot. Automat. Lett.}, vol. 7, no. 2, pp. 5238–5245, 2022.

\bibitem{c34} S. Kumra, S. Joshi, and F. Sahin, “Antipodal robotic grasping using
generative residual convolutional neural network,” in \emph{Proc. IEEE/RSJ Int. Conf. Intell. Robots Syst.}, 2020, pp. 9626–9633.

\bibitem{c35} M. Shan, J. Zhang, H. Zhu, C. Li, and F. Tian, "Grasp Detection Algorithm Based on CPS-ResNet," in \emph{Proc. IEEE Int. Conf. Image Process. Comput. Vis. Mach. Learn.}, 2022, pp. 501-506.

\bibitem{c36} H. Cao, G. Chen, Z. Li, Q. Feng, J. Lin, and A. Knoll, “Efficient grasp detection network with Gaussian-based grasp representation for robotic manipulation,”\emph{IEEE/ASME Trans. Mech.}, vol. 28, no. 3, pp. 1384–1394, 2022.

\bibitem{c37} S. Yu, D.-H. Zhai, and Y. Xia, “CGNet: Robotic grasp detection in heavily cluttered scenes,” \emph{IEEE/ASME Trans. Mech.}, vol. 28, no. 2, pp. 884–894, 2023.

\bibitem{c38} J. Mahler et al., “Dex-Net 1.0: A cloud-based network of 3D objects for robust grasp planning using a multi-armed bandit model with correlated rewards,” in \emph{Proc. IEEE Int. Conf. Robot. Automat.}, 2016, pp. 1957–1964.

\bibitem{c39} J. Mahler et al., “Dex-Net 2.0: Deep learning to plan robust grasps with synthetic point clouds and analytic grasp metrics,” in \emph{Proc. Robot.: Sci. Syst.}, 2017.

\bibitem{c40} J. Mahler, M. Matl, X. Liu, A. Li, D. Gealy, and K. Goldberg, “Dex-Net 3.0: Computing robust vacuum suction grasp targets in point clouds using a new analytic model and deep learning,” in \emph{Proc. IEEE Int. Conf. Robot. Automat.}, 2018, pp. 5620–5627.

\bibitem{c41} J. Mahler and K. Goldberg, “Learning deep policies for robot bin picking by simulating robust grasping sequences,” in \emph{Conf. Robot Learn.}, 2017, pp. 515–524.

\bibitem{c42} H. S. Fang, C. Wang, M. Gou, and C. Lu, “GraspNet-1billion: A large scale benchmark for general object grasping,” in \emph{Proc. IEEE Conf. Comput. Vis. Pattern Recognit.}, 2020, pp. 11444–11453.

\bibitem{c43} H. S. Fang et al., “AnyGrasp: Robust and efficient grasp perception in spatial and temporal domains,” \emph{IEEE Trans. Robot.}, vol. 39, no. 5, pp. 3929–3945, 2023.

\bibitem{c44} C. Szegedy et al., “Intriguing properties of neural networks,” in \emph{Proc. Int. Conf. Learn. Representations.}, 2014.

\bibitem{c45} J. Wang, “Adversarial examples in physical world,” in \emph{Proc. Int. Joint Conf. Artif. Intell.}, 2021, pp. 4925–4926.

\bibitem{c46} X. Liu, H. Yang, Z. Liu, L. Song, Y. Chen, and H. Li, “DPATCH: An adversarial patch attack on object detectors,” in \emph{Proc. AAAI Conf. Artif. Intell.}, 2019, pp. 1–8.

\bibitem{c47} M. Lee and Z. Kolter, “On physical adversarial patches for object detection,” 2019, \emph{arXiv:} 1906.11897.

\bibitem{c48} Y. Jiang, S. Moseson, and A. Saxena, “Efficient grasping from RGBD images: Learning using a new rectangle representation,” in \emph{Proc. IEEE Int. Conf. Robot. Automat.}, 2011, pp. 3304–3311.

\bibitem{c49} D. P. Kingma and J. Ba, “Adam: A method for stochastic optimization,” 2014, \emph{arXiv:} 1412.6980.

\bibitem{c50} M. Gruosso, N. Capece, and U. Erra, “Egocentric upper limb segmentation in unconstrained real-life scenarios“,  \emph{Virtual Reality.}, vol. 27, pp. 3421–3433, 2023.

\bibitem{c51} C. Li, Z. Gao, and N. Y. Chong, "Shortcut-enhanced Multimodal Backdoor Attack in Vision-guided Robot Grasping," \emph{Authorea Preprints.}, 2024.

\bibitem{c52} A. Depierre, E. Dellandréa, and L. Chen, “Jacquard: A large scale dataset for robotic grasp detection,” in \emph{Proc. IEEE/RSJ Int. Conf. Intell. Robots Syst.}, 2018, pp. 3511–3516.

\bibitem{c53}  S. Ainetter and F. Fraundorfer, “End-to-end trainable deep neural network for robotic grasp detection and semantic segmentation from RGB,” in \emph{Proc. IEEE Int. Conf. Robot. Automat.}, 2021, pp. 13452–13458.

\bibitem{c54} M. Suchi, T. Patten, and M. Vincze, “EasyLabel: A semi-automatic pixel-wise object annotation tool for creating robotic RGB-D datasets,” in \emph{Proc. IEEE Conf. Robot. Automat.}, 2019, pp. 6678–6684.

\bibitem{c55} S. Wang, Z. Zhou, and Z. Kan, “When transformer meets robotic grasping: Exploits context for efficient grasp detection,” \emph{IEEE Robot. Automat. Lett.}, vol. 7, no. 3, pp. 8170–8177, 2022.
\end{thebibliography}
\end{document}